\ifdefined\XeTeXversion\else
  \ifdefined\pdfoutput\pdfoutput=1\relax\fi
  \ifdefined\pdfminorversion\pdfminorversion=7\relax\fi
\fi
\documentclass{article}
\usepackage[T1]{fontenc}

\usepackage{microtype}
\usepackage{booktabs}
\usepackage{hyperref}
\usepackage{amsmath}
\usepackage{amsthm}
\usepackage{amssymb}
\usepackage{xspace}
\usepackage{tabularx}
\usepackage[svgnames]{xcolor}
\usepackage{multirow}
\usepackage{colortbl}
\usepackage{subcaption}

\makeatletter
\RequirePackage{times}

\RequirePackage{fancyhdr}
\RequirePackage{color}
\RequirePackage{algorithm}
\RequirePackage{algorithmic}
\RequirePackage{natbib}
\RequirePackage{eso-pic} \RequirePackage{forloop}

\ifdefined\nohyperref\else\ifdefined\hypersetup
  \definecolor{mydarkblue}{rgb}{0,0.08,0.45}
  \hypersetup{ pdftitle={},
    pdfauthor={},
    pdfsubject={},
    pdfkeywords={},
    pdfborder=0 0 0,
    pdfpagemode=UseNone,
    colorlinks=true,
    linkcolor=mydarkblue,
    citecolor=mydarkblue,
    filecolor=mydarkblue,
    urlcolor=mydarkblue,
    pdfview=FitH}

  \ifdefined\isaccepted \else
    \hypersetup{pdfauthor={Anonymous Submission}}
  \fi
\fi\fi

\newcommand{\Notice@String}{Preprint.}

\gdef\isaccepted{1}

\renewcommand{\cite}[1]{\citep{#1}}

\newif\ifpdf \ifx\pdfoutput\undefined
\else
   \ifx\pdfoutput\relax
   \else
     \ifcase\pdfoutput
     \else
       \pdftrue
     \fi
   \fi
\fi

\ifpdf
\fi

\AtBeginDocument{\newif\ifmarginsmessedwith
\marginsmessedwithfalse
\ifdim\oddsidemargin=-16.62178pt     \else oddsidemargin has been altered.\\ \marginsmessedwithtrue\fi
\ifdim\headheight=10.0pt             \else headheight has been altered.\\ \marginsmessedwithtrue\fi
\ifdim\textheight=650.43pt           \else textheight has been altered.\\ \marginsmessedwithtrue\fi
\ifdim\marginparsep=11.0pt           \else marginparsep has been altered.\\ \marginsmessedwithtrue\fi
\ifdim\footskip=0.0pt                \else footskip has been altered.\\ \marginsmessedwithtrue\fi
\ifdim\hoffset=0.0pt                 \else hoffset has been altered.\\ \marginsmessedwithtrue\fi
\ifdim\paperwidth=614.295pt          \else paperwidth has been altered.\\ \marginsmessedwithtrue\fi
\ifdim\topmargin=-24.95781pt         \else topmargin has been altered.\\ \marginsmessedwithtrue\fi
\ifdim\headsep=10.0pt                \else headsep has been altered.\\ \marginsmessedwithtrue\fi
\ifdim\textwidth=487.8225pt          \else textwidth has been altered.\\ \marginsmessedwithtrue\fi
\ifdim\marginparwidth=65.0pt         \else marginparwidth has been altered.\\ \marginsmessedwithtrue\fi
\ifdim\marginparpush=5.0pt           \else marginparpush has been altered.\\ \marginsmessedwithtrue\fi
\ifdim\voffset=0.0pt                 \else voffset has been altered.\\ \marginsmessedwithtrue\fi
\ifdim\paperheight=794.96999pt       \else paperheight has been altered.\\ \marginsmessedwithtrue\fi
\ifmarginsmessedwith

\textbf{\large \em The page layout violates the PAPER style.}

Please do not change the page layout, or include packages like geometry,
savetrees, or fullpage, which change it for you.

We're not able to reliably undo arbitrary changes to the style. Please remove
the offending package(s), or layout-changing commands and try again.

\fi}

\def\ftype@copyrightbox{8}
\def\@copyrightspace{
\@float{copyrightbox}[b]
\begin{center}
\setlength{\unitlength}{1pc}
\begin{picture}(20,1.5)
\put(0,2.5){\line(1,0){4.818}}
\put(0,0){\parbox[b]{19.75pc}{\small \Notice@String}}
\end{picture}
\end{center}
\end@float}

 \twocolumn
\def\addcontentsline#1#2#3{}

\newbox\titrun

\def\papertitlerunning#1{\gdef\@papertitlerunning{#1}}

\long\def\papertitle#1{

   \ifx\undefined\@papertitlerunning\gdef\@papertitlerunning{#1}
   \fi

  \ifdefined\nohyperref\else\ifdefined\hypersetup
     \hypersetup{pdftitle={#1}}
   \fi\fi

   \global\setbox\titrun=\vbox{\small\bf\@papertitlerunning}

   \gdef\@runningtitleerror{0}

   \ifdim\wd\titrun>\textwidth{\gdef\@runningtitleerror{1}}\else\ifdim\ht\titrun>6.25pt
	   {\gdef\@runningtitleerror{2}}\fi
   \fi

   \ifnum\@runningtitleerror>0
	   \typeout{}\typeout{}\typeout{*******************************************************}\typeout{Title exceeds size limitations for running head.}\typeout{Please supply a shorter form for the running head}
           \typeout{with \string\papertitlerunning{...}\space prior to \string\begin{document}}\typeout{*******************************************************}\typeout{}\typeout{}\chead{\small\bf Title Suppressed Due to Excessive Size}\else
  	   \chead{\small\bf\@papertitlerunning}\fi

  \thispagestyle{empty}

  {\center\baselineskip 18pt
                       \toptitlebar{\Large\bf\scshape #1}\bottomtitlebar}
}

\gdef\paperfullauthorlist{}
\newcommand\addstringtofullauthorlist{\g@addto@macro\paperfullauthorlist}
\newcommand\addtofullauthorlist[1]{\ifdefined\paperanyauthors\addstringtofullauthorlist{, #1}\else\addstringtofullauthorlist{#1}\gdef\paperanyauthors{1}\fi\ifdefined\nohyperref\else\ifdefined\hypersetup\hypersetup{pdfauthor=\paperfullauthorlist}\fi\fi}

\def\toptitlebar{\hrule height2pt \vskip .25in}
\def\bottomtitlebar{\vskip .22in \hrule height1pt \vskip .3in}

\newenvironment{paperauthorlist}{\setlength\topsep{0pt}
  \setlength\parskip{0pt}
  \begin{center}
}{\end{center}
}

\newcounter{@affiliationcounter}
\newcommand{\@pa}[1]{\ifcsname the@affil#1\endcsname
\else
  \ifcsname @papersymbol#1\endcsname
  \else
  \stepcounter{@affiliationcounter}\newcounter{@affil#1}\setcounter{@affil#1}{\value{@affiliationcounter}}\fi
\fi\ifcsname @papersymbol#1\endcsname
  \textsuperscript{\csname @papersymbol#1\endcsname\,}\else
  \textsuperscript{\arabic{@affil#1}\,}\fi
}

\newcommand{\paperauthor}[2]{\ifdefined\isaccepted
    \mbox{\bf #1}\,\@for\theaffil:=#2\do{\@pa{\theaffil}} \addtofullauthorlist{#1}\else
    \ifdefined\@paperfirsttime
    \else
      \gdef\@paperfirsttime{1}
      \mbox{\bf Anonymous Authors}\@pa{@anon} \addtofullauthorlist{Anonymous Authors}
     \fi
    \fi
}

\newcommand{\papersetsymbol}[2]{\expandafter\gdef\csname @papersymbol#1\endcsname{#2}
 }

\newcommand{\paperaffiliation}[2]{\ifdefined\isaccepted
\ifcsname the@affil#1\endcsname
 \expandafter\gdef\csname @affilname\csname the@affil#1\endcsname\endcsname{#2}\else
  {\bf AUTHORERR: Error in use of \textbackslash{}paperaffiliation command. Label ``#1'' not mentioned in some \textbackslash{}paperauthor\{author name\}\{labels here\} command beforehand. }
  \typeout{}\typeout{}\typeout{*******************************************************}\typeout{Affiliation label undefined. }\typeout{Make sure \string\paperaffiliation\space follows }
  \typeout{all of \string\paperauthor\space commands}\typeout{*******************************************************}\typeout{}\typeout{}\fi
\else \expandafter\gdef\csname @affilname1\endcsname{Anonymous Institution, Anonymous City, Anonymous Region, Anonymous Country}
\fi
}

\newcommand{\papercorrespondingauthor}[2]{
\ifdefined\isaccepted
 \ifdefined\papercorrespondingauthor@text
   \g@addto@macro\papercorrespondingauthor@text{, #1 \textless{}#2\textgreater{}}
 \else
   \gdef\papercorrespondingauthor@text{#1 \textless{}#2\textgreater{}}
 \fi
\else
\gdef\papercorrespondingauthor@text{Anonymous Author \textless{}anon.email@domain.com\textgreater{}}
\fi
}

\newcommand{\paperEqualContribution}{\textsuperscript{*}Equal contribution }

\newcounter{@affilnum}
\newcommand{\printAffiliationsAndNotice}[1]{\stepcounter{@affiliationcounter}{\let\thefootnote\relax\footnotetext{\hspace*{-\footnotesep}\ifdefined\isaccepted #1\fi\forloop{@affilnum}{1}{\value{@affilnum} < \value{@affiliationcounter}}{
\textsuperscript{\arabic{@affilnum}}\ifcsname @affilname\the@affilnum\endcsname\csname @affilname\the@affilnum\endcsname\else
{\bf AUTHORERR: Missing \textbackslash{}paperaffiliation.}
\fi
}.
\ifdefined\papercorrespondingauthor@text
Correspondence to: \papercorrespondingauthor@text.
\else
{\bf AUTHORERR: Missing \textbackslash{}papercorrespondingauthor.}
\fi

\ \\
\Notice@String
}
}
}

\long\def\paperaddress#1{{\bf The \textbackslash{}paperaddress command is no longer used.  See the example\_paper PDF .tex for usage of \textbackslash{}paperauther and \textbackslash{}paperaffiliation.}
}

\def\paperkeywords#1{\ifdefined\nohyperref\else\ifdefined\hypersetup
    \hypersetup{pdfkeywords={#1}}
  \fi\fi
}

\setcitestyle{authoryear,round,citesep={;},aysep={,},yysep={;}}

\renewenvironment{abstract}
   {\centerline{\large\bf\scshape Abstract}
   \vspace{-0.12in}\begin{quote}}
   {\par\end{quote}\vskip 0.12in}

\def\@startsection#1#2#3#4#5#6{\if@noskipsec \leavevmode \fi
   \par \@tempskipa #4\relax
   \@afterindenttrue
   \ifdim \@tempskipa <\z@ \@tempskipa -\@tempskipa \fi
   \if@nobreak \everypar{}\else
     \addpenalty{\@secpenalty}\addvspace{\@tempskipa}\fi \@ifstar
     {\@ssect{#3}{#4}{#5}{#6}}{\@dblarg{\@sict{#1}{#2}{#3}{#4}{#5}{#6}}}}

\def\@sict#1#2#3#4#5#6[#7]#8{\ifnum #2>\c@secnumdepth
     \def\@svsec{}\else
     \refstepcounter{#1}\edef\@svsec{\csname the#1\endcsname}\fi
     \@tempskipa #5\relax
      \ifdim \@tempskipa>\z@
        \begingroup #6\relax
          \@hangfrom{\hskip #3\relax\@svsec\quad}{\interlinepenalty \@M #8\par}
        \endgroup
       \csname #1mark\endcsname{#7}\addcontentsline
         {toc}{#1}{\ifnum #2>\c@secnumdepth \else
                      \protect\numberline{\csname the#1\endcsname}\fi
                    #7}\else
        \def\@svsechd{#6\hskip #3\@svsec #8\csname #1mark\endcsname
                      {#7}\addcontentsline
                           {toc}{#1}{\ifnum #2>\c@secnumdepth \else
                             \protect\numberline{\csname the#1\endcsname}\fi
                       #7}}\fi
     \@xsect{#5}}

\def\@sect#1#2#3#4#5#6[#7]#8{\ifnum #2>\c@secnumdepth
     \def\@svsec{}\else
     \refstepcounter{#1}\edef\@svsec{\csname the#1\endcsname\hskip 0.4em }\fi
     \@tempskipa #5\relax
      \ifdim \@tempskipa>\z@
        \begingroup #6\relax
          \@hangfrom{\hskip #3\relax\@svsec}{\interlinepenalty \@M #8\par}
        \endgroup
       \csname #1mark\endcsname{#7}\addcontentsline
         {toc}{#1}{\ifnum #2>\c@secnumdepth \else
                      \protect\numberline{\csname the#1\endcsname}\fi
                    #7}\else
        \def\@svsechd{#6\hskip #3\@svsec #8\csname #1mark\endcsname
                      {#7}\addcontentsline
                           {toc}{#1}{\ifnum #2>\c@secnumdepth \else
                             \protect\numberline{\csname the#1\endcsname}\fi
                       #7}}\fi
     \@xsect{#5}}

\def\thesection {\arabic{section}}
\def\thesubsection {\thesection.\arabic{subsection}}
\def\section{\@startsection{section}{1}{\z@}{-0.12in}{0.02in}
             {\large\bf\scshape\raggedright}}
\def\subsection{\@startsection{subsection}{2}{\z@}{-0.10in}{0.01in}
                {\normalsize\bf\raggedright}}
\def\subsubsection{\@startsection{subsubsection}{3}{\z@}{-0.08in}{0.01in}
                {\normalsize\itshape\raggedright}}
\def\paragraph{\@startsection{paragraph}{4}{\z@}{1.5ex plus
  0.5ex minus .2ex}{-1em}{\normalsize\bf}}
\def\subparagraph{\@startsection{subparagraph}{5}{\z@}{1.5ex plus
  0.5ex minus .2ex}{-1em}{\normalsize\bf}}

\def\footnoterule{\kern-3pt \hrule width 0.8in \kern 2.6pt }
\labelwidth\leftmargini\advance\labelwidth-\labelsep \labelsep 5pt

\def\@listi{\leftmargin\leftmargini}
\def\@listii{\leftmargin\leftmarginii
   \labelwidth\leftmarginii\advance\labelwidth-\labelsep
   \topsep 2pt plus 1pt minus 0.5pt
   \parsep 1pt plus 0.5pt minus 0.5pt
   \itemsep \parsep}
\def\@listiii{\leftmargin\leftmarginiii
    \labelwidth\leftmarginiii\advance\labelwidth-\labelsep
    \topsep 1pt plus 0.5pt minus 0.5pt
    \parsep \z@ \partopsep 0.5pt plus 0pt minus 0.5pt
    \itemsep \topsep}
\def\@listiv{\leftmargin\leftmarginiv
     \labelwidth\leftmarginiv\advance\labelwidth-\labelsep}
\def\@listv{\leftmargin\leftmarginv
     \labelwidth\leftmarginv\advance\labelwidth-\labelsep}
\def\@listvi{\leftmargin\leftmarginvi
     \labelwidth\leftmarginvi\advance\labelwidth-\labelsep}

\def\@normalsize{\@setsize\normalsize{11pt}\xpt\@xpt}
\def\small{\@setsize\small{10pt}\ixpt\@ixpt}
\def\footnotesize{\@setsize\footnotesize{10pt}\ixpt\@ixpt}
\def\scriptsize{\@setsize\scriptsize{8pt}\viipt\@viipt}
\def\tiny{\@setsize\tiny{7pt}\vipt\@vipt}
\def\large{\@setsize\large{14pt}\xiipt\@xiipt}
\def\Large{\@setsize\Large{16pt}\xivpt\@xivpt}
\def\LARGE{\@setsize\LARGE{20pt}\xviipt\@xviipt}
\def\huge{\@setsize\huge{23pt}\xxpt\@xxpt}
\def\Huge{\@setsize\Huge{28pt}\xxvpt\@xxvpt}

\newsavebox\newcaptionbox\newdimen\newcaptionboxwid

\long\def\@makecaption#1#2{
 \vskip 10pt
        \baselineskip 11pt
        \setbox\@tempboxa\hbox{#1. #2}
        \ifdim \wd\@tempboxa >\hsize
        \sbox{\newcaptionbox}{\small\sl #1.~}
        \newcaptionboxwid=\wd\newcaptionbox
        \usebox\newcaptionbox {\footnotesize #2}
        \else
          \centerline{{\small\sl #1.} {\small #2}}
        \fi}

\def\fnum@figure{Figure \thefigure}
\def\fnum@table{Table \thetable}

\def\abovestrut#1{\rule[0in]{0in}{#1}\ignorespaces}
\def\belowstrut#1{\rule[-#1]{0in}{#1}\ignorespaces}

\def\abovespace{\abovestrut{0.20in}}
\def\aroundspace{\abovestrut{0.20in}\belowstrut{0.10in}}
\def\belowspace{\belowstrut{0.10in}}

\def\texitem#1{\par\noindent\hangindent 12pt
               \hbox to 12pt {\hss #1 ~}\ignorespaces}
\def\paperitem{\texitem{$\bullet$}}

\makeatletter
\newbox\paperrulerbox
\newcount\paperrulercount
\newdimen\paperruleroffset
\newdimen\cv@lineheight
\newdimen\cv@boxheight
\newbox\cv@tmpbox
\newcount\cv@refno
\newcount\cv@tot
\newcount\cv@tmpc@ \newcount\cv@tmpc
\def\fillzeros[#1]#2{\cv@tmpc@=#2\relax\ifnum\cv@tmpc@<0\cv@tmpc@=-\cv@tmpc@\fi
\cv@tmpc=1 \loop\ifnum\cv@tmpc@<10 \else \divide\cv@tmpc@ by 10 \advance\cv@tmpc by 1 \fi
   \ifnum\cv@tmpc@=10\relax\cv@tmpc@=11\relax\fi \ifnum\cv@tmpc@>10 \repeat
\ifnum#2<0\advance\cv@tmpc1\relax-\fi
\loop\ifnum\cv@tmpc<#1\relax0\advance\cv@tmpc1\relax\fi \ifnum\cv@tmpc<#1 \repeat
\cv@tmpc@=#2\relax\ifnum\cv@tmpc@<0\cv@tmpc@=-\cv@tmpc@\fi \relax\the\cv@tmpc@}\def\makevruler[#1][#2][#3][#4][#5]{
	\begingroup\offinterlineskip
		\textheight=#5\vbadness=10000\vfuzz=120ex\overfullrule=0pt\global\setbox\paperrulerbox=\vbox to \textheight{{
				\parskip=0pt\hfuzz=150em\cv@boxheight=\textheight
				\cv@lineheight=#1\global\paperrulercount=#2\cv@tot\cv@boxheight\divide\cv@tot\cv@lineheight\advance\cv@tot2\cv@refno1\vskip-\cv@lineheight\vskip1ex\loop\setbox\cv@tmpbox=\hbox to0cm{					 \hfil {\hfil\fillzeros[#4]\paperrulercount}
				}\ht\cv@tmpbox\cv@lineheight\dp\cv@tmpbox0pt\box\cv@tmpbox\break
				\advance\cv@refno1\global\advance\paperrulercount#3\relax
				\ifnum\cv@refno<\cv@tot\repeat
			}
		}
	\endgroup
}\makeatother

\def\paperruler#1{\makevruler[12pt][#1][1][3][\textheight]\usebox{\paperrulerbox}}
\AddToShipoutPicture{\paperruleroffset=\textheight
\advance\paperruleroffset by 5.2pt \color[rgb]{.7,.7,.7}
  \ifdefined\isaccepted \else
	  \AtTextUpperLeft{\put(\LenToUnit{-35pt},\LenToUnit{-\paperruleroffset}){\paperruler{\paperrulercount}}
	  }
	 \fi
}

\makeatother

\makeatletter
\renewcommand{\Notice@String}{Preprint. Copyright 2026 by the author(s).}
\makeatother

\usepackage{graphicx}

\usepackage[nostruts]{titlesec}
\titlespacing*{\section}{0pt}{0pt}{0pt}
\titlespacing*{\subsection}{0pt}{0pt}{0pt}

\usepackage{float}
\newtheorem{theorem}{Theorem}[section]
\newtheorem{corollary}[theorem]{Corollary}

\usepackage{enumitem}
\setlist[itemize]{nosep, leftmargin=*}
\setlist[enumerate]{nosep, leftmargin=*}

\newcommand{\name}{Spool\xspace}

\newcommand{\TTFTSpeedup}{$6.6\times$\xspace}
\newcommand{\OnlineGain}{2.3 percentage points\xspace}

\begin{document}

\twocolumn[
  \papertitle{\name: Low-Latency Streaming Inference for Hybrid Vision-Language Models}

  \begin{paperauthorlist}
    \paperauthor{Rui Ning}{csail}
    \paperauthor{Om Chabra}{csail}
    \paperauthor{Hari Balakrishnan}{csail}
  \end{paperauthorlist}

  \paperaffiliation{csail}{Computer Science and Artificial Intelligence Laboratory, Massachusetts Institute of Technology, Cambridge, MA, USA}

  \papercorrespondingauthor{Hari Balakrishnan}{hari@csail.mit.edu}

  \paperkeywords{}

  \vskip 0.3in

\begin{abstract}
  Applications such as visual accessibility agents and robotics require answering questions about recent video content promptly, over a bounded window of past frames (e.g., the last two minutes). Hybrid vision-language models (VLMs), which interleave attention with linear recurrent layers such as Gated DeltaNet (GDN), are attractive for this setting, but maintaining a sliding window is expensive. The reason is that recurrent layers combine all past frames into a shared state, so expiring old frames normally requires re-running the entire window through the model. We observe that the recurrent update is {\em affine} in the state with input-dependent coefficients, so the effect of each frame can be captured as a transformation that composes associatively. We present \name, a training-free framework that caches per-frame state transformations and composes them efficiently as the window advances, removing expired frames from each recurrent layer's state without replaying frames through the backbone. Removal is exact within each layer; as with key-value eviction in attention, expired frames can retain residual influence through the cached inputs computed by earlier layers. We design a dedicated data structure and GPU kernels to achieve low caching and composition overhead. Across five open-source hybrid VLMs and four video reasoning benchmarks, \name achieves the lowest p95 time-to-first-token in all 20 model--benchmark tests, reducing it by a geometric mean of \TTFTSpeedup over the faster of two recent vLLM baselines, while improving accuracy by \OnlineGain on average.
\end{abstract}

]

\makeatletter
\chead{\small\bf\@papertitlerunning}
\makeatother

\printAffiliationsAndNotice{}

\section{Introduction}
\label{sec:intro}

\begin{figure*}[!htbp]
  \centering
  \includegraphics[width=0.99\textwidth]{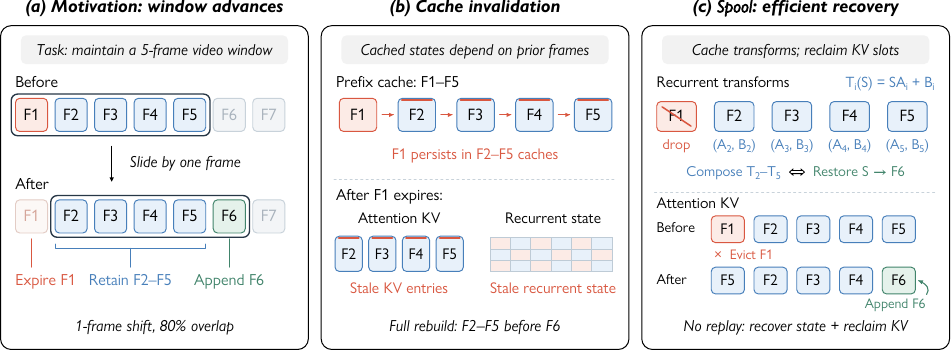}
  \caption{\textbf{\name maintains a sliding video window without replaying retained frames.} (a) Shifting the window retains four of five frames. (b) Prefix-cache invalidation forces the retained context to be rebuilt. (c) \name composes cached recurrent transformations and reclaims key--value slots by moving surviving tail entries into holes left by eviction.}
  \label{fig:overview}
\end{figure*}

In streaming video understanding, a vision-language model (VLM) continuously ingests video frames and answers user prompts in real time. Consider the example where a robot receives a request and reasons through successive actions using its camera. Responses must be fast and accurate because the environment changes rapidly, new frames arrive continually, and poor visual perception leads to inaccurate actions. Other applications include visual accessibility agents~\cite{icanseeforever}, medical video understanding~\cite{medgrpo}, interactive live sports analytics~\cite{livecc, streamingvlm}, streaming content moderation~\cite{livestream-moderation}, and industrial and manufacturing applications~\cite{factory-vad}.

The key challenge is that these models (e.g., Qwen3.5-VL~\cite{qwen35}, Nemotron 3.5 Super VL \cite{nemotron35}) incur computational cost for every new frame and eventually reach the model's maximum context length. Streaming systems therefore maintain a sliding window: they keep the video's last $N$ frames and evict older ones (e.g., with $N=60$, frame $1$ is evicted when frame $61$ arrives). Serving engines such as vLLM~\cite{pagedattention} and SGLang~\cite{sglang} reuse computation across requests through prefix caching. However, prefix caching requires the beginning of the sequence to remain unchanged. Once the window evicts the first frame, the cache is invalidated, forcing recomputation of every remaining frame (e.g., frames 2 through 61).

Attention layers can avoid this recomputation by evicting the key-value (KV) entries of expired frames directly. Hybrid VLMs, however, make the problem more difficult. They interleave two types of layers (Fig.~\ref{fig:architecture}): attention layers, which store per-token KV entries for retrieving past information, and linear recurrent layers such as Gated DeltaNet (GDN) \cite{gdn}, which mix successive frames into a single fixed-size state. KV entries can be evicted individually, but the recurrent state has no per-frame entries to remove. For queries restricted to the current window, expired frames should no longer affect the answer, yet their influence persists in the recurrent state after the corresponding KV entries are evicted. Removing it normally requires re-running the entire window through the model, a slow process.

To address this problem, we introduce \textbf{\name}. Our key insight is that, with no additional training, \emph{we can cache how each video frame transforms the recurrent state, and reuse these transformations as the window advances}. This lets \name both incorporate new frames and remove the influence of expired frames without re-running the window.

\name addresses three challenges.

\textbf{Challenge 1: Removing expired frames from recurrent state.} Unlike attention, which keeps a memory entry per token, a recurrent layer keeps a fixed-size state that every new frame updates (Fig.~\ref{fig:overview}). Conventional state checkpointing (i.e., storing the GDN state after every frame) does not help, because each checkpoint already contains information from earlier frames that should expire. Our observation is that \emph{a GDN (or Mamba) update is an affine transformation of the previous state}, $S_t = S_{t-1}A_t + B_t$, so the updates for all tokens in a frame compose into a single affine transformation. Rather than caching states, \name caches the multiplicative and additive coefficients $(A, B)$ of each frame's transformation. Composing the coefficients of the frames in the current window and applying the result to the initial state rebuilds each recurrent layer's state without re-running the backbone. Removal is exact within each layer; expired frames retain residual influence only through the inputs that earlier layers computed while those frames were in context, as with KV eviction in attention. We show in \S\ref{sec:evaluation} that this residual effect has little impact on accuracy, and that caching the coefficients requires only a modest amount of extra memory (4.8\,GiB, or 18\%, for a 9B model).

\textbf{Challenge 2: Making reconstruction efficient.} Applying the cached transformation of every frame in the window takes $O(W)$ work per update, where $W$ is the window length in frames. Because affine transformations compose associatively, \name caches the combined transformations of consecutive frames in a segment tree and reconstructs the state with $O(\log W)$ compositions.

\textbf{Challenge 3: Reclaiming attention memory.} Unlike the recurrent state, attention KV entries can be removed individually. However, serving engines store the KV cache in fixed-size pages. Evicting early frames leaves partially occupied pages, and compacting the surviving entries would require copying memory proportional to the retained context. \name instead fills vacated KV slots with the entries of newly arriving frames. To prevent positional indices from growing indefinitely, \name periodically rebases cached rotary position embeddings (RoPE), preserving relative-position attention without recomputing visual representations.

We evaluate \name on six open-source hybrid VLMs, from 0.8B to 122B parameters with GDN and Mamba-2 recurrent layers, and four streaming video benchmarks, against vLLM-Omni with and without its recently added incremental prefill. In online serving, \name achieves the lowest p95 time-to-first-token in all 20 model--benchmark tests, reducing it by a geometric mean of \TTFTSpeedup (up to $35.7\times$) over the faster baseline and $33.8\times$ over vLLM-Omni, keeping p95 under 1\,s in 16 of 20 model--benchmark tests compared with 3 for incremental prefill. Because fewer questions are dropped under streaming load, it also improves online accuracy by \OnlineGain on average and by up to 18.4 points on Beyond3D. Without timing pressure, \name stays within $-1.6$ to $+1.8$ points of the more accurate baseline in all 24 offline tests. Its caching components reduce per-frame ingestion latency by 7--17\% relative to a naive cache, KV filling saves up to 1.8\,GiB of peak memory, and after 1024 expired frames, state reconstruction keeps KL divergence from vLLM-Omni about $2.5\times$ lower than without it.

\section{Background and Motivation}
\label{sec:motivation}

\subsection{VLMs for Realtime Video Understanding}

VLMs combine visual perception with language understanding, allowing users to ask natural language questions about images and video. For video, they can reason across frames about events and how a scene changes over time.

Streaming applications (\S\ref{sec:intro}) impose three requirements. First, answers must arrive promptly and reflect recent observations; second, processing must keep pace with incoming frames; and third, services handling many concurrent streams need high throughput at low cost. These requirements are hard to meet because every frame requires visual encoding and language model computation.

A streaming VLM processes frames incrementally and retains model state to reuse earlier computation. When a question arrives, it generates an answer from the accumulated video context and the question. To operate within finite resources on long-running streams, the system keeps a sliding window of recent frames, in which older frames expire as new ones arrive. This workflow involves three operations: video updates, question answering, and maintaining the state for the active window.

\subsection{Hybrid VLM Architecture}

\begin{figure}
  \centering
  \includegraphics[width=\columnwidth]{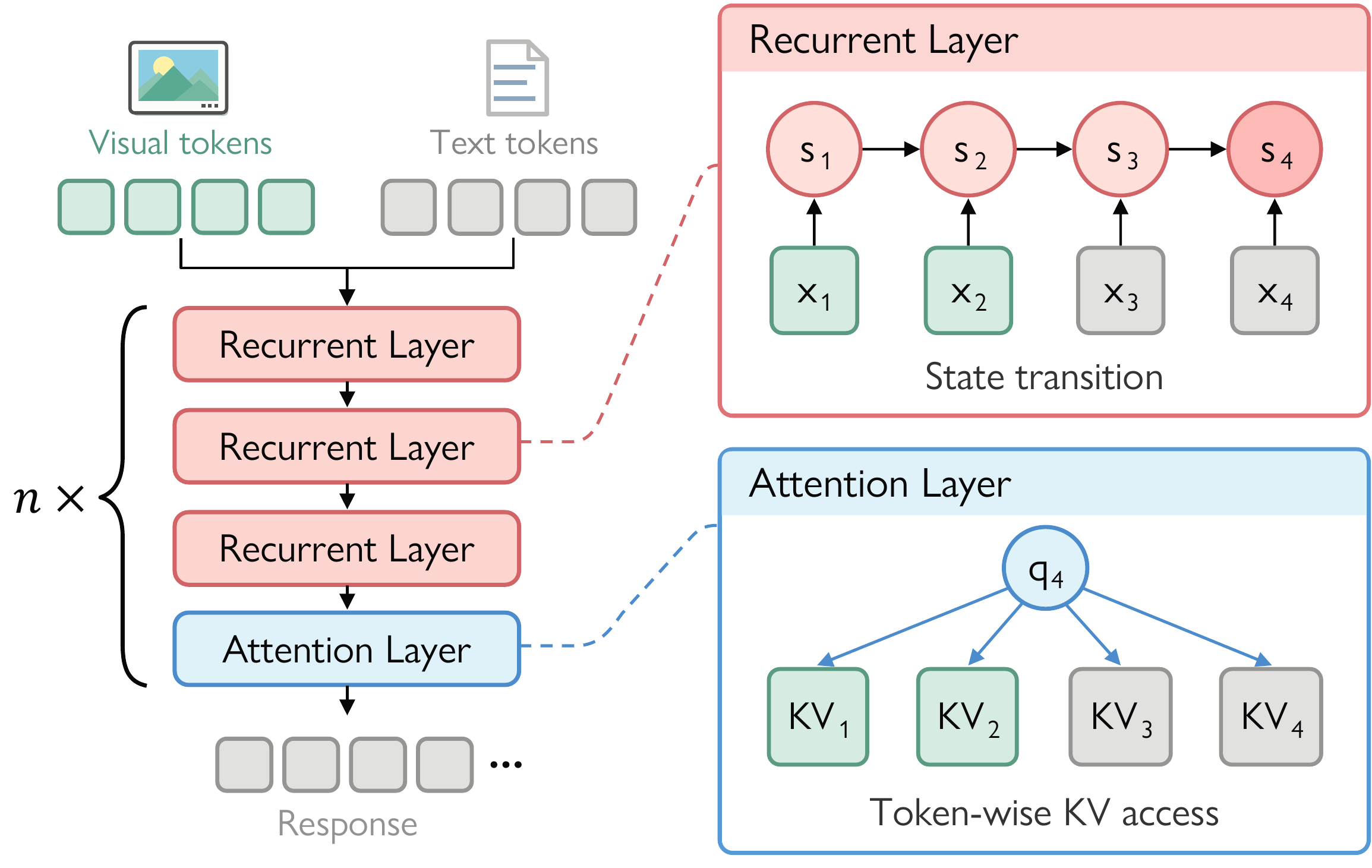}
  \caption{\textbf{Architecture of the hybrid VLM.} \textnormal{Multiple input frames are converted into visual tokens by a vision encoder and passed alongside text tokens into a unified backbone. This backbone interleaves recurrent layers, which update a fixed-size state for each token, with attention layers, which retain per-token key--value entries.}}
  \label{fig:architecture}
\end{figure}

Figure~\ref{fig:architecture} illustrates the architecture of a hybrid VLM. A vision encoder converts each incoming frame into a sequence of visual tokens, and instructions and questions are embedded as text tokens. The language backbone processes the combined sequence causally and generates answer tokens autoregressively.

The backbone interleaves two types of sequence-mixing layers that differ in how they represent history: attention layers retain separate entries for earlier tokens, while recurrent layers summarize history in a fixed-size state.

\textbf{Attention layers.} An attention layer accesses preceding tokens through their key and value vectors, which are retained in a KV cache during inference. New tokens reuse these cached entries to compute attention outputs. The cache preserves separately addressable entries for individual tokens, allowing the model to precisely retrieve specific pieces of past information, though the required memory size increases directly with the length of the context~\cite{pagedattention}.

\textbf{Recurrent layers.} Recurrent layers, including Mamba-style state-space layers and Gated DeltaNet~(GDN), accumulate history in a fixed-size state. This allows them to process long sequence lengths by completely decoupling memory consumption from the amount of context observed. They help a model maintain a persistent summary of past information~\cite{mamba,mamba2,gdn}.

For fixed layer inputs, their core updates take the affine form
\begin{equation}
  s_t = A_t s_{t-1} + b_t,
\end{equation}
where $A_t$ transforms existing memory and $b_t$ incorporates new information. State memory remains independent of sequence length, with earlier inputs combined within the shared state.

\subsection{KV Cache Management}

\paragraph{Paged storage.} The KV cache grows as new tokens are processed. Paged attention divides its storage into fixed-size chunks of GPU memory, called \emph{pages}, each with space for a fixed number of tokens' key--value pairs~\cite{pagedattention}. A sequence of tokens/blocks/frame entries can span multiple pages, which need not be adjacent in memory. A \emph{page table} records which page holds each consecutive group of entries, and the attention computation consults this table to find the keys and values it needs. The cache can therefore grow by allocating additional pages as needed.

Because frames produce variable numbers of tokens, frame boundaries often fall across KV pages. As a simple example, consider a page of four token entries, where the first six tokens belong to expired frames. Evicting these tokens frees the first page (tokens 1--4), but the second page must retain tokens 7--8. This leaves two invalid entries (tokens 5--6) alongside live entries in the same page. These \emph{holes} cannot be reclaimed by freeing the page and must be excluded from attention.

\paragraph{Position information.} A token's logical position identifies where it occurs in the input sequence; its storage location identifies where its KV entry resides in memory. An attention layer compares a new token's query vector with cached keys to compute weights for a weighted sum of their associated values. Rotary position embeddings (RoPE) encode token positions by rotating query and key vectors according to position indices~\cite{roformer}. The resulting attention scores depend on relative position differences. Since keys are cached after rotation, each stored key already carries position information. Evicting old tokens reduces the number of cached entries, but does not automatically reset the positions assigned to subsequent tokens.

\subsection{Context Window Maintenance}

Streaming VLMs continuously process a live stream of video blocks while answering intermittent user questions. Each block contains multiple contiguous frames, and each frame can produce a different number of visual tokens. To respect the model's context limit, the system maintains a fixed prefix (e.g., the system prompt) followed by a \emph{sliding window} of recent video blocks. The window is constrained by both frame and token budgets, with tokens reserved for questions and generated responses.

Consider a window of five blocks (Fig.~\ref{fig:overview}). As frames $F_1$ through $F_5$ arrive, the model incrementally appends them, using prefix caching to reuse computed states. However, at maximum capacity, the arrival of $F_6$ requires evicting $F_1$, shifting the window from

\[
    [F_1,F_2,F_3,F_4,F_5]
    \;\longrightarrow\;
    [F_2,F_3,F_4,F_5,F_6].
\]

Although four blocks remain unchanged, their cached recurrent representations were computed with $F_1$ in the context, making its information remain mixed into a shared state. Consequently, evicting $F_1$ leaves stale information that can interfere with answers about the active window, particularly when expired and current content conflict (\citealp{streamingllm,stuffedmamba}; see also \S\ref{sec:evaluation}).

\begin{figure}[!htbp]
  \centering
  \includegraphics[width=0.8\linewidth]{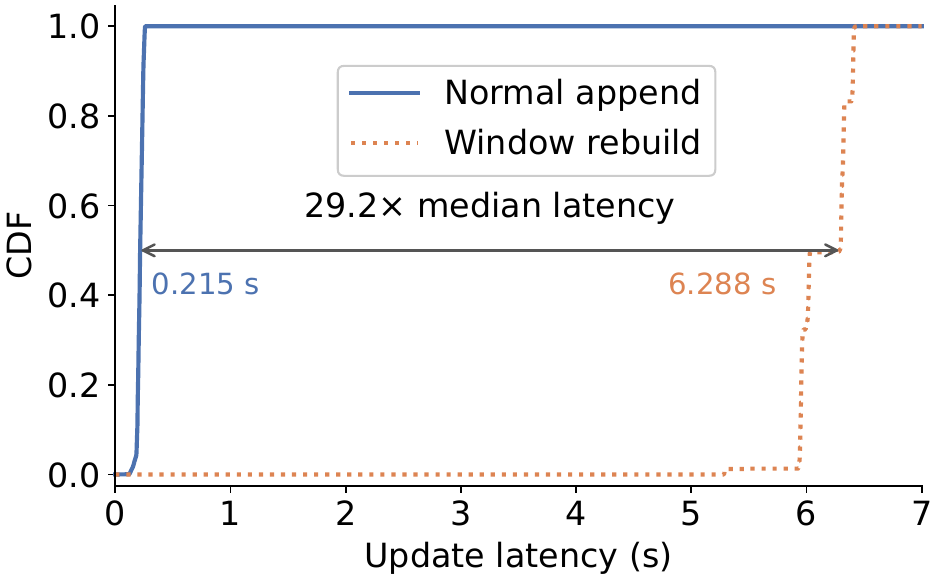}
  \caption{\textbf{Cost of window rebuilding.} Here we plot time for a frame to be processed by Qwen3.5-9B on 1 A100. Every time the window must be rebuilt from scratch, a 29.2x latency cost must be paid. }
  \label{fig:window-rebuild}
\end{figure}

Conventional implementations must therefore recompute the entirety of the retained blocks when inserting a frame ($F_6$). This \emph{window rebuild} repeats computation for existing blocks in the context window $(F_2,\ldots, F_5)$, despite their substantial overlap with the previous window. When evaluated with a Qwen3.5-9B (NVIDIA A100, 128 frame window), we observe that a full rebuild takes 6.288\,s at the median, compared with 0.215\,s for a normal block append---a $29.2\times$ increase (Fig.~\ref{fig:window-rebuild}). Such latency spikes are particularly costly for real-time video understanding.

Recent approaches like vLLM-Omni's incremental-prefill strategy~\cite{vllmomni} work by reducing the rebuild frequency by evicting in batches rather than advancing the window one block at a time. In our evaluated configuration, it discards the oldest half of the window when capacity is reached. This approach has three limitations: (1) full rebuilds still cause tail-latency spikes; (2) batch eviction discards potentially useful context that could otherwise remain within the budget; and (3) reserving capacity for batch eviction prevents continuous use of the full context budget.

This highlights an opportunity to eliminate redundancy by reusing the cached states of retained blocks. The central challenge, however, is reusing these states while effectively removing the influence of blocks no longer in the active context.

\section{Design}
\label{sec:design}

\name is a training-free inference engine that efficiently maintains a bounded video context window for VLMs. \name centers around using internal state of previously processed processed frames. In particular, \name has two complementary mechanisms: (1) \textbf{recurrent state reconstruction}, which caches transformations of previously processed video blocks to maintain accurate recurrent state, and  (2) \textbf{attention cache reclamation}, which minimizes data movement for expiring KV entries.

We first describe \name's execution model (\S\ref{sec:streaming-execution}), followed by recurrent state reconstruction (\S\ref{sec:streaming-execution}), its efficient implementation (\S\ref{sec:summary-tree}), and attention cache reclamation (\S\ref{sec:windowed-attention-state}).

\subsection{Streaming Execution Model}
\label{sec:streaming-execution}

\noindent\textbf{Active Window.} \name represents the active video window as a block range $[a,b)$. When admitting a new block that would exceed either budget, the engine solely advances $a$ to evict the oldest blocks. Eviction always occurs at whole-block boundaries, even though individual blocks can contain different numbers of frames and tokens. Once a video is at the set context length budget, all future frames will mantain the same length through eviction.

\noindent\textbf{Session State.} The engine prefills the fixed system prompt once and saves its recurrent state as an anchor for subsequent reconstruction. The engine then maintains three types of model state: (1) recurrent states, (2) cached recurrent transformations, and (3) attention KV pages. The engine additionally tracks metadata including token identities, block boundaries, and positions to keep the internal data consistent with the main active window.

\noindent\textbf{Video Block Updates.} When a new video block arrives, the engine first determines which existing blocks must expire. If eviction is necessary, it removes the expired blocks' KV entries and reconstructs the recurrent state of the retained window using the stored block transitions, as described in \S\ref{sec:windowed-recurrent-state} and \S\ref{sec:windowed-attention-state}. The engine then prefills the incoming block, records its recurrent transition, and updates the segment tree. If no eviction is required, prefill continues directly from the live state. In either case, the engine updates the context window without reprocessing the retained video tokens, leaving it ready for the next block or question.

\subsection{Maintaining Windowed Recurrent State}
\label{sec:windowed-recurrent-state}

As established in \S\ref{sec:streaming-execution}, failing to remove expired blocks from the recurrent state degrades model accuracy, while reprocessing all retained tokens to rebuild it is prohibitively expensive. \name solves this problem by caching each video block's state transformation. As we show below, recurrent layers such as Gated Delta Net (GDN) and State Space Model update their state with an affine map. Each block's effect is a transition $F_i(S) = S A_i + B_i$ that applies to any incoming state. \name rebuilds this state by replaying cached blocks' transitions, without processing any tokens again ($A_i$ or $B_i$).

The rest of this section explains why the recurrent updates follow this form, describes how \name records and applies these transitions, and discusses the limitations of the approach.

\paragraph{Recurrent updates are affine.} Consider one head of a recurrent layer, with state $S \in \mathbb{R}^{d_v \times d_k}$. Here we use GDN as the running example, but the same procedure applies independently to multiple recurrent layers and heads. We will first consider the update step for one token $t$ which then expands to one video block $i$.

At token $t$, the layer uses its input $x_t$ to compute four quantities: a normalized key $k_t \in \mathbb{R}^{d_k}$, a value $v_t \in \mathbb{R}^{d_v}$, a decay gate $\alpha_t$, and a write gate $\beta_t$. The update first decays the memory by $\alpha_t$. It then erases the entry at $k_t$ and writes $v_t$ in its place, with strength $\beta_t$:
\begin{equation}
  \begin{aligned}
    S_t &= \alpha_t S_{t-1} + \beta_t \bigl(v_t - \alpha_t S_{t-1} k_t\bigr) k_t^{\mathsf T} \\
    &= S_{t-1} A_t + B_t \\
    A_t &= \alpha_t \bigl(I - \beta_t k_t k_t^{\mathsf T}\bigr) \\
    B_t &= \beta_t v_t k_t^{\mathsf T}
  \end{aligned}
  \label{eq:gdn-affine}
\end{equation}

Importantly, the update of $S_t$ is affine because $k_t$, $v_t$, $\alpha_t$, and $\beta_t$ are computed only from $x_t$ and the state $S_{t-1}$ is not an input to any of them, with $A_t$ and $B_t$ depending solely on token $t$.

Now, considering multiple tokens, this form remains preserved, because the composition of two affine maps gives another affine map. Substituting the update for token $t$ into the update for token $t+1$ shows this directly:
\begin{align}
  S_{t+1} &= \bigl(S_{t-1} A_t + B_t\bigr) A_{t+1} + B_{t+1}
  \label{eq:combinedtransition} \\
  &= S_{t-1} \bigl(A_t A_{t+1}\bigr)
  + \bigl(B_t A_{t+1} + B_{t+1}\bigr) \nonumber
\end{align}

The result is again a matrix product plus an offset: the matrices multiply together, and the earlier write $B_t$ is carried forward through the next token's matrix. Repeating this substitution across every token of video block $i$ collapses the block into a single transition:
\begin{equation}
  F_i(S) = S A_i + B_i, \qquad A_i \in \mathbb{R}^{d_k \times d_k},\; B_i \in \mathbb{R}^{d_v \times d_k}
  \label{eq:block-transition}
\end{equation}
Here, $A_i$ is the ordered product of the block's token matrices, and $B_i$ accumulates the block's writes. Because neither depends on the state entering the block, \name can record $A_i$ and $B_i$ once and later apply it to a different incoming state. Other recurrent layers like SSM share this structure, with a scalar or diagonal decay in place of $A_t$, so the same construction carries over.

\paragraph{Recording and applying block transitions.}

The derivation above guarantees that every block has a transition $(A, B)$; the remaining question is how does \name obtain and use these transitions? As a reminder, \name maintains a working context window of blocks $[a, b)$. As each video block (i.e. $b$) enters the window, \name caches its transition as a by-product of normal prefill (Figure~\ref{fig:recurrent-reconstruction}, top). It builds $A_i$ from intermediates that the recurrent kernel already computes (\S\ref{sec:implementation}). \name recovers $B_i$ from the states on either side of the block (i.e., by Eq.~\eqref{eq:block-transition}, $  B_i = S_{\mathrm{out}} - S_{\mathrm{in}} A_i$).

During normal streaming, the VLM will compute the default transition matrices, while \name will cache each block's transition. The VLLM will continue to use the block's default output state $S_{\mathrm{out}}$ as the live state to prefill the next block.

When a given block expires, \name will use the stored transitions to reconstruct the state without the expired block. It starts from the anchor state $S_{\mathrm p}$, saved after the prefill of the fixed prefix or prompt, and will recompute the expected state using the transitions stored.

As an example, suppose the window holds blocks $[a, b)$, so \name has cached the transitions $F_a, \dots, F_{b-1}$. When block $b$ arrives, and the window is full, the oldest block $a$ must expire. \name discards $(A_a, B_a)$ and rebuilds the recurrent state of the remaining blocks $[a+1, b)$ by applying their transitions, in chronological order, to a precomputed anchor $S_{\mathrm p}$:

\begin{equation}
  S \leftarrow S A_j + B_j \quad \text{for } j = a+1, \dots, b-1
  \label{eq:reconstruction}
\end{equation}
The final value of $S$ is the reconstructed state $\widehat S_{a+1,b}$. Block $a$ simply drops out: $(A_a, B_a)$ is never applied, and none of the retained blocks' tokens pass through the backbone again.

\paragraph{Limitations of this approach} Within each recurrent layer, transition caching and replay are lossless with respect to the recorded state updates. As the inputs to the GDN (the keys, values, and gates in Eq. \ref{eq:gdn-affine}) are held fixed, each cached pair $(A_i, B_i)$ exactly represents the updates in block $i$ to any layer's recurrent state. Replaying the retained transitions from the layer's prefix anchor $S_{\mathrm p}$ therefore yields the same state as executing those recorded updates as the default process.

However, this layerwise exactness does not imply equivalence to rerunning the full network. In a multilayer network, the input features to a given recurrent layer are produced by preceding layers and may already reflect blocks that have since expired. Consequently, the recorded coefficients $A_i$ and $B_i$ can depend on the history present when block $i$ was originally processed. Omitting expired transitions therefore removes their direct contributions to a given layer, but does not eliminate their influence already encoded in the retained transitions. As we show in \S\ref{sec:evaluation} and Appendix~\ref{app:gdn-state-error}, relative to full-network recomputation, \name introduces only a small, bounded approximation error that does not accumulate as the window advances.

\smallskip
\noindent\textbf{Residual summary representation.} To limit the memory overhead of caching block transition matrices, \name uses bfloat16 by default. However, $A$ is often close to the identity matrix (see Eq.~\ref{eq:gdn-affine}), so directly storing it in low precision can round away small changes to its diagonal entries. To mitigate this, \name instead stores the residual $D=A-I$, explicitly separating the identity component from the small deviations, better preserved in low precision, with the original state transition evaluated as $S+SD+B$.

\subsection{Maintaining Recurrent State Efficiently}
\label{sec:summary-tree}

\begin{figure}[!t]
  \centering
  \includegraphics[width=\columnwidth]{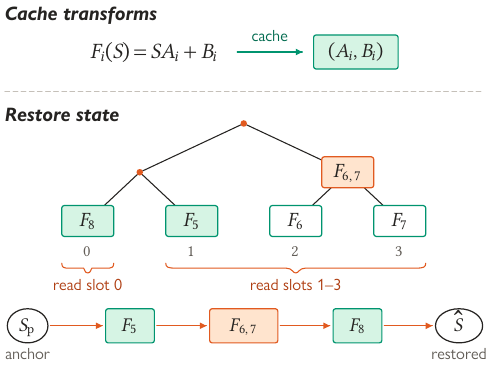}
\caption{\textbf{Efficient recurrent-state reconstruction with cached transformations.}
Spool caches each block's affine state transformation and then replays them without the effect of a given block. A segment tree composes consecutive transformations, enabling
reconstruction of the active window in $O(\log W)$ operations.
In this example, frames 5--8 occupy cyclic slots 1--3 and 0,
with frame 8 wrapping around to slot 0.
Spool first applies $F_5$, then the combined transformation
$F_{6,7}$, and finally $F_8$ to the prefix state $S_p$.
This recovers the window state using only three cached
transformations, without reprocessing video tokens or updating
the tree during reconstruction.}  \label{fig:recurrent-reconstruction}
\vspace{5mm}
\end{figure}

To avoid sequentially applying the transitions for the entire context window each time a block expires, \name caches combined transformations of consecutive video blocks in a segment tree (Fig. \ref{fig:recurrent-reconstruction}). By exploiting the associative property of affine transformations, this reduces recurrent state reconstruction from $O(W)$ sequential block applications to $O(\log W)$, where $W$ is the window length in blocks.

As we show below, a sequence of block transformations follows the same affine representation as a single transformation. This allows any block-aligned interval to be hierarchically decomposed into tree nodes whose summaries are applied in chronological order.

\smallskip
\noindent\textbf{Composition of affine summaries.} We first show that consecutive block transformations can be represented by a single affine transformation. For a block interval $[a,b)$, let its combined transformation be $F_{a,b}(S)=SA_{a,b}+B_{a,b}$.

For any split point $a\leq k\leq b$, applying the transformations of $[a,k)$ and $[k,b)$ consecutively gives
\begin{equation}
  \begin{aligned}
    F_{a,b}(S) &= F_{k,b}\bigl(F_{a,k}(S)\bigr),\\
    A_{a,b} &= A_{a,k}A_{k,b},\\
    B_{a,b} &= B_{a,k}A_{k,b}+B_{k,b}.
  \end{aligned}
  \label{eq:summary-composition}
\end{equation}

Thus, composing consecutive affine transformations produces another affine transformation of the same form (Eq.   \ref{eq:combinedtransition}). Since function composition is associative, these summaries can be merged hierarchically without changing the represented transformation, provided their chronological order is preserved. Formally, affine summaries form a non-commutative monoid under Equation~\eqref{eq:summary-composition}, with identity $(I,0)$.

Additionally, the residual summary representation introduced above still preserves this associative property. Substituting $A=I+D$ into Equation~\eqref{eq:summary-composition} gives the composition of two consecutive residual summaries:
\begin{equation}
  \begin{aligned}
    D_{a,b} &= D_{a,k}+D_{k,b}+D_{a,k}D_{k,b},\\
    B_{a,b} &= B_{a,k}+B_{a,k}D_{k,b}+B_{k,b}.
  \end{aligned}
  \label{eq:residual-composition}
\end{equation}

This allows the segment tree to compose residual summaries directly without changing the underlying interval operations, with proofs provided in Appendix~\ref{app:summary-monoids}.

\smallskip
\noindent\textbf{Segment-tree structure.} We exploit this associative property using a segment tree, where each leaf stores the transformation $F_i$ of a video block, and each internal node stores the composition of its two children. Specifically, a node representing $[a,b)$ stores $F_{a,b}$, obtained by composing $F_{a,k}$ and $F_{k,b}$ from its left and right children (Fig. \ref{fig:recurrent-reconstruction}).

\smallskip
\noindent\textbf{Insertion and tree updates.} \name assigns a given block to $W$ cyclically reused leaf slots. Each computed block's recorded transformation is stored in the next available leaf slot. It then updates all ancestors of that leaf, composing their children's summaries bottom-up while maintaining order. Each insertion therefore requires $O(\log W)$ summary compositions. If a video block spans multiple prefill batches, its summaries are composed into a single leaf summary.

\smallskip
\noindent\textbf{Eviction and state reconstruction.} When eviction is required, the engine first advances the window boundary past all expired blocks and reconstructs the recurrent state from the remaining blocks (often 1 or 2 operations). Since the leaf slots are reused cyclically, the blocks may wrap around the circular layout, occupying at most two contiguous ranges of leaf slots. Expired blocks are excluded from the reconstruction process, and their leaf slots can be overwritten by subsequent insertions.

Fig. \ref{fig:recurrent-reconstruction} illustrates this process with frames 5--8 stored across four cyclic slots. Because the slots are reused cyclically, frame 8 occupies slot 0, while frames 5--7 occupy slots 1--3. To reconstruct the state, the engine processes the slots in chronological order: first frame 5 (slot 1), then the combined summary for frames 6--7 (slots 2--3), and finally frame 8 (slot 0). Each cached transformation updates the state using $S \leftarrow SA+B$. Thus, the engine reconstructs the state using just three cached transformations, without reprocessing the frames, updating the tree, or constructing a new summary for the entire window.

\smallskip
\noindent\textbf{Computational and memory cost.} To mantain the segment tree, \name requires updating the tree on each append, while the full state reconstruction occurs only upon eviction.

For each recurrent layer and value head, combining two cached transformations requires $O(d_k^3+d_vd_k^2)$ operations. Since inserting a new block updates $O(\log W)$ ancestor nodes in the segment tree, each insertion costs $O((d_k^3+d_vd_k^2)\log W)$. When a block expires, Spool reconstructs the recurrent state by applying the $O(\log W)$ cached transformations covering the retained window to the prefix state, at a total cost of $O(d_vd_k^2\log W)$.

The segment tree uses additional memory to store cached transformations at each node. For $L_g$ recurrent layers with $H_v$ value heads each, every node stores a transformation $(A,B)$ per head, where $A\in\mathbb{R}^{d_k\times d_k}$ and $B\in\mathbb{R}^{d_v\times d_k}$. Assuming uniform head dimensions and $s$ bytes per scalar, a tree padded to $P=2^{\lceil\log_2 W\rceil}$ leaves requires the following memory in bytes:

\begin{equation}
  \mathcal{M}_{\mathrm{tree}} = 2PL_gH_v(d_k^2+d_vd_k)s.
\end{equation}

\subsection{Maintaining Windowed Attention State}
\label{sec:windowed-attention-state}

In addition to maintaining the recurrent state, \name must efficiently update the attention key-value (KV) cache as video blocks expire.
Unlike recurrent layers, attention layers store individual token entries in fixed-size memory pages.
However, video blocks contain varying numbers of tokens, so expiring a block often leaves partially occupied KV pages that waste GPU memory or require a compaction step that introduces additional latency.
A second challenge is that attention assigns each new token a position index, which continues to increase as new video blocks arrive, even after older tokens expire. Over a long-running stream, these indices eventually exceed the model's precomputed rotary position embedding (RoPE) table, which supports only a fixed range of positions.

\name addresses these challenges with two mechanisms: (1) \textbf{KV tail filling}, which reclaims memory by moving surviving KV entries from the end of the cache into slots left by expired tokens, avoiding copies of the entire retained context; and (2) \textbf{RoPE rebasing}, which periodically shifts token positions and rotates cached keys to preserve attention outputs without recomputing retained tokens.

\subsubsection{KV tail filling}

To reclaim expired KV slots without shifting existing tokens, \name uses tail filling to keep the cache dense. Importantly, existing KV entries do not have to maintain their original order because each key encodes its position. Therefore, assuming the keys and values are moved together, reordering the KV pairs therefore preserves the attention output in exact arithmetic. \name exploits this property to fill expired slots with surviving KV pairs from the cache tail.

\noindent\textbf{Filling Expired Slots.}
To identify expired tokens after previous relocations, \name maintains a mapping between tokens and their current KV slots. Suppose the cache contains $n$ tokens (including a fixed prefix) and the window advancement expires $e$ video tokens. The engine keeps the prefix unchanged and compacts the remaining KV entries into the first $n-e$ slots. For each expired slot in this range, it moves a surviving KV pair from the tail into the hole and updates the mapping (Fig.~\ref{fig:kv-tail-filling}, top). Expired entries already in the tail require no movement.

\noindent\textbf{Copy Cost.}
Tail filling copies only the surviving KV pairs needed to fill expired slots. Let $m$ be the number of relocated tokens and $b_{\mathrm{KV}}$ the total KV payload per token across attention layers. Each relocation reads and writes one KV pair, giving a total copy cost of $2mb_{\mathrm{KV}}$ bytes. Since each relocation fills one expired slot, $m \le e$. Thus, the copy cost depends on the number of expired tokens rather than the length of the retained window.

\noindent\textbf{Page Reclamation.}
After filling the expired slots, the engine reduces the sequence length to $n-e$ and releases pages entirely beyond this. Any partially filled final page remains allocated for subsequent tokens. The resulting cache is dense and can be used by existing prefill and decode kernels, without modifying the attention kernels.

\subsubsection{RoPE rebasing}

\begin{figure}[!t]
  \centering
  \includegraphics[width=\columnwidth]{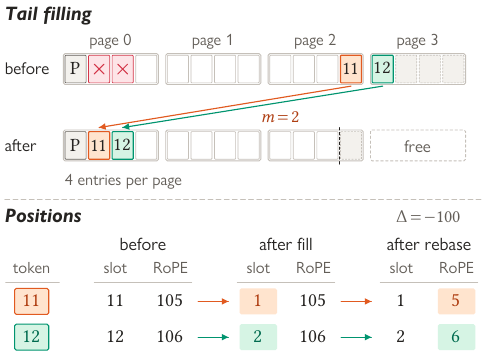}
  \caption{\textbf{KV tail filling and RoPE rebasing.} Top: surviving tokens 11 and 12 fill holes in slots 1 and 2, releasing the final page; $\mathrm{P}$ denotes the fixed prefix. Bottom: filling changes storage slots while preserving position coordinates. A separate rebase shifts all cached positions by a common offset ($-100$ here) and rotates cached keys consistently.}
  \label{fig:kv-tail-filling}
\end{figure}

To prevent token positions from exceeding the precomputed RoPE table, \name periodically shifts the position coordinates of the context window. Although eviction limits the number of cached tokens, the position offset continues to increase as new video blocks arrive. \name addresses this by subtracting a common offset from token positions and adjusting the cached keys, without reprocessing retained tokens (Fig.~\ref{fig:kv-tail-filling}, bottom).

\noindent\textbf{Preserving Attention Scores.}
RoPE encodes token positions by rotating query and key vectors. Consider a query $q_i$ at position $p_i$ and a key $k_j$ at position $p_j$, before applying RoPE. Let $\mathcal{R}(p)$ denote the rotation corresponding to position $p$. RoPE produces the rotated vectors $\widetilde q_i$ and $\widetilde k_j$, whose attention score is
\begin{equation}
  \begin{aligned}
    \widetilde q_i &= \mathcal{R}(p_i)q_i \\
    \widetilde k_j &= \mathcal{R}(p_j)k_j \\
    \widetilde q_i^{\mathsf T}\widetilde k_j
    &= q_i^{\mathsf T}\mathcal{R}(p_j-p_i)k_j.
  \end{aligned}
\end{equation}
The final equality follows from the property
$\mathcal{R}(u)^{\mathsf T}\mathcal{R}(v)=\mathcal{R}(v-u)$.
Importantly, the attention score depends on the relative positions of the query and key, rather than their absolute positions. Therefore, subtracting the same offset $\Delta$ from both positions preserves their attention score.

However, changing the position coordinates alone is insufficient because cached keys have already been rotated using their original positions. To account for the shift, \name applies an additional rotation $\mathcal{R}(-\Delta)$ to each cached key. Since RoPE rotations compose additively, this produces the same key as applying RoPE at the shifted position:
\begin{equation}
  \begin{aligned}
    p'_j &= p_j-\Delta \\
    \widetilde k'_j
    &= \mathcal{R}(-\Delta)\widetilde k_j \\
    &= \mathcal{R}(p_j-\Delta)k_j.
  \end{aligned}
\end{equation}
Thus, \name can update the position encoded in each cached key without recomputing its original representation.

New queries use the shifted position $p'_i=p_i-\Delta$, producing $\widetilde q'_i=\mathcal{R}(p_i-\Delta)q_i$. Because both queries and keys use the same shifted coordinate system, the common offset cancels when computing their attention score:
\begin{equation}
  \begin{aligned}
    (\widetilde q'_i)^{\mathsf T}\widetilde k'_j
    &= q_i^{\mathsf T}
       \mathcal{R}\bigl((p_j-\Delta)-(p_i-\Delta)\bigr)k_j \\
    &= q_i^{\mathsf T}\mathcal{R}(p_j-p_i)k_j \\
    &= \widetilde q_i^{\mathsf T}\widetilde k_j.
  \end{aligned}
\end{equation}
Thus, rebasing preserves attention scores and, with the cached values and attention mask unchanged, also preserves attention outputs in exact arithmetic.

\noindent\textbf{Rebasing Cached Keys.}
Before an incoming block or query would exceed the RoPE table, \name subtracts $\Delta$ from the stored position coordinates and the next position offset. It then applies $\mathcal{R}(-\Delta)$ to all cached keys, including the fixed prefix. For multi-axis RoPE, \name applies the same shift to the time, height, and width coordinates, with the corresponding rotations applied independently to each rotary subspace. This allows the engine to continue processing new tokens without extending the RoPE table or reprocessing the retained context.

\section{Implementation}
\label{sec:implementation}

We implement \name in Python with PyTorch, Triton, and Flash Linear Attention, targeting hybrid VLMs. The runtime loads pretrained checkpoints and uses their Hugging Face processors for frame preparation and prompt construction.

\textbf{Recurrent kernels.} We extend Flash Linear Attention's chunked GDN prefill path to record block summaries from its existing intermediates. Custom Triton kernels construct residual summaries, merge tree nodes, and apply summaries to recurrent states. Merges and reconstruction batch independent layers and heads. Following the vLLM precision defaults, summaries use BF16 storage by default, while the prefix anchor and live recurrent state remain in FP32. State-transition products split the FP32 state into high and low BF16 components and accumulate both products in FP32.

\textbf{Attention cache.} The cache stores BF16 keys and values in 256-token pages, using PyTorch's paged variable-length attention for prefill and Triton kernels for decoding. For tail filling, the host translates the relocation plan through the page table into physical source and destination slots. A single Triton grid copies the selected entries across all attention layers and both keys and values. A separate kernel rebases cached keys using FP32 arithmetic before storing them in BF16.

\section{Evaluation}
\label{sec:evaluation}

\begin{table*}[!t]
\centering\small
\setlength{\tabcolsep}{1pt}
\renewcommand{\arraystretch}{1.18}
\begin{tabularx}{\linewidth}{@{}cc*{5}{>{\centering\arraybackslash}X}@{}}
  \toprule
  \multirow{2}{*}{\textbf{Benchmark}} & \multirow{2}{*}{\textbf{Method}} & \textbf{Qwen3.5} & \mbox{\textbf{Nemotron-VL}} & \textbf{Qwen3.5} & \textbf{Qwen3.6} & \textbf{Qwen3.5} \\
  & & \textbf{9B} & \textbf{Nano-12B} & \textbf{27B} & \textbf{35B-A3B} & \textbf{122B-A10B} \\
  \midrule
  \multirow{3}{*}{\textbf{PhoStream}} & vLLM-Omni & 31.53 & 5.52 & 12.11 & 8.13 & 5.57 \\
  & vLLM-Incremental & 12.22 & 2.23 & 5.79 & 2.44 & 2.02 \\
  \rowcolor{LightSteelBlue!20}\cellcolor{white} & \textcolor{SteelBlue!50!black}{\textbf{\name (Ours)}} & \textbf{0.34}\,{\scriptsize$(\times35.7)$} & \textbf{0.17}\,{\scriptsize$(\times13.0)$} & \textbf{0.29}\,{\scriptsize$(\times20.0)$} & \textbf{0.91}\,{\scriptsize$(\times2.7)$} & \textbf{1.31}\,{\scriptsize$(\times1.5)$} \\
  \addlinespace[5pt]
  \multirow{3}{*}{\textbf{StreamingBench}} & vLLM-Omni & 31.59 & 5.55 & 9.64 & 11.74 & 8.02 \\
  & vLLM-Incremental & 11.70 & 2.24 & 4.51 & 1.84 & 1.99 \\
  \rowcolor{LightSteelBlue!20}\cellcolor{white} & \textcolor{SteelBlue!50!black}{\textbf{\name (Ours)}} & \textbf{0.35}\,{\scriptsize$(\times33.8)$} & \textbf{0.17}\,{\scriptsize$(\times13.3)$} & \textbf{0.28}\,{\scriptsize$(\times16.2)$} & \textbf{0.25}\,{\scriptsize$(\times7.4)$} & \textbf{0.33}\,{\scriptsize$(\times6.0)$} \\
  \addlinespace[5pt]
  \multirow{3}{*}{\textbf{FastBench}} & vLLM-Omni & 4.00 & 6.00 & 5.23 & 0.88 & 1.04 \\
  & vLLM-Incremental & 0.90 & 4.03 & 2.71 & 0.31 & 0.45 \\
  \rowcolor{LightSteelBlue!20}\cellcolor{white} & \textcolor{SteelBlue!50!black}{\textbf{\name (Ours)}} & \textbf{0.13}\,{\scriptsize$(\times6.8)$} & \textbf{0.14}\,{\scriptsize$(\times28.2)$} & \textbf{0.23}\,{\scriptsize$(\times11.9)$} & \textbf{0.15}\,{\scriptsize$(\times2.1)$} & \textbf{0.21}\,{\scriptsize$(\times2.1)$} \\
  \addlinespace[5pt]
  \multirow{3}{*}{\textbf{BEYOND3D}} & vLLM-Omni & 60.55 & 191.42 & 298.61 & 287.89 & 258.73 \\
  & vLLM-Incremental & 2.33 & 6.96 & 6.17 & 8.03 & 9.09 \\
  \rowcolor{LightSteelBlue!20}\cellcolor{white} & \textcolor{SteelBlue!50!black}{\textbf{\name (Ours)}} & \textbf{0.81}\,{\scriptsize$(\times2.9)$} & \textbf{0.97}\,{\scriptsize$(\times7.2)$} & \textbf{1.82}\,{\scriptsize$(\times3.4)$} & \textbf{3.49}\,{\scriptsize$(\times2.3)$} & \textbf{6.60}\,{\scriptsize$(\times1.4)$} \\
  \bottomrule
\end{tabularx}
\caption{\textbf{Online p95 TTFT (s, $\downarrow$).} \name can maintain real-time performance as live video streams in, achieving the lowest tail latency in all 20 model--benchmark tests: a geometric-mean speedup of \TTFTSpeedup over the faster baseline and $33.8\times$ over vLLM-Omni. Gains are largest on smaller models (0.34\,s vs.\ 12.22\,s on PhoStream with Qwen3.5-9B, $35.7\times$) and remain clear at the largest scale (0.33\,s vs.\ 1.99\,s on StreamingBench with Qwen3.5-122B-A10B). (Parentheses are speedup over the faster baseline.)}
\label{tab:online-p95}
\end{table*}

\begin{table*}[!t]
\centering\small
\setlength{\tabcolsep}{1pt}
\renewcommand{\arraystretch}{1.18}
\begin{tabularx}{\linewidth}{@{}cc*{5}{>{\centering\arraybackslash}X}@{}}
  \toprule
  \multirow{2}{*}{\textbf{Benchmark}} & \multirow{2}{*}{\textbf{Method}} & \textbf{Qwen3.5} & \mbox{\textbf{Nemotron-VL}} & \textbf{Qwen3.5} & \textbf{Qwen3.6} & \textbf{Qwen3.5} \\
  & & \textbf{9B} & \textbf{Nano-12B} & \textbf{27B} & \textbf{35B-A3B} & \textbf{122B-A10B} \\
  \midrule
  \multirow{3}{*}{\textbf{PhoStream}} & vLLM-Omni & 54.06 & 55.72 & 67.16 & 67.66 & 70.15 \\
  & vLLM-Incremental & 61.25 & \textbf{57.71} & \textbf{71.64} & \textbf{69.65} & \textbf{71.14} \\
  \rowcolor{LightSteelBlue!20}\cellcolor{white} & \textcolor{SteelBlue!50!black}{\textbf{\name (Ours)}} & \textbf{64.27}\,{\scriptsize$(+3.0\%)$} & 56.22\,{\scriptsize$(-1.5\%)$} & 69.15\,{\scriptsize$(-2.5\%)$} & 69.15\,{\scriptsize$(-0.5\%)$} & \textbf{71.14}\,{\scriptsize$(+0.0\%)$} \\
  \addlinespace[5pt]
  \multirow{3}{*}{\textbf{StreamingBench}} & vLLM-Omni & 78.79 & 72.50 & 80.63 & 80.00 & 85.63 \\
  & vLLM-Incremental & \textbf{84.24} & \textbf{77.50} & \textbf{86.88} & \textbf{83.75} & \textbf{88.13} \\
  \rowcolor{LightSteelBlue!20}\cellcolor{white} & \textcolor{SteelBlue!50!black}{\textbf{\name (Ours)}} & \textbf{84.24}\,{\scriptsize$(+0.0\%)$} & 75.00\,{\scriptsize$(-2.5\%)$} & \textbf{86.88}\,{\scriptsize$(+0.0\%)$} & 82.50\,{\scriptsize$(-1.3\%)$} & 86.88\,{\scriptsize$(-1.3\%)$} \\
  \addlinespace[5pt]
  \multirow{3}{*}{\textbf{FastBench}} & vLLM-Omni & 32.35 & 30.39 & 40.20 & 34.31 & \textbf{37.91} \\
  & vLLM-Incremental & 31.05 & 29.41 & 39.22 & \textbf{35.95} & 35.95 \\
  \rowcolor{LightSteelBlue!20}\cellcolor{white} & \textcolor{SteelBlue!50!black}{\textbf{\name (Ours)}} & \textbf{35.95}\,{\scriptsize$(+3.6\%)$} & \textbf{30.72}\,{\scriptsize$(+0.3\%)$} & \textbf{41.83}\,{\scriptsize$(+1.6\%)$} & 35.29\,{\scriptsize$(-0.7\%)$} & 37.58\,{\scriptsize$(-0.3\%)$} \\
  \addlinespace[5pt]
  \multirow{3}{*}{\textbf{BEYOND3D}} & vLLM-Omni & 6.78 & 3.44 & 4.21 & 2.54 & 2.64 \\
  & vLLM-Incremental & 22.93 & 11.94 & 24.74 & 12.40 & 10.08 \\
  \rowcolor{LightSteelBlue!20}\cellcolor{white} & \textcolor{SteelBlue!50!black}{\textbf{\name (Ours)}} & \textbf{31.41}\,{\scriptsize$(+8.5\%)$} & \textbf{30.32}\,{\scriptsize$(+18.4\%)$} & \textbf{35.14}\,{\scriptsize$(+10.4\%)$} & \textbf{19.04}\,{\scriptsize$(+6.6\%)$} & \textbf{13.71}\,{\scriptsize$(+3.6\%)$} \\
  \bottomrule
\end{tabularx}
\caption{\textbf{Online accuracy (\%, $\uparrow$).} By answering on time, \name drops fewer questions under live load and answers more correctly than both baselines where they fall behind. On BEYOND3D, it improves over the more accurate baseline by 3.6--18.4 points (mean 9.5), such as 30.32\% vs.\ 11.94\% with Nemotron-VL Nano-12B. On the other benchmarks, it stays within $-2.5$ to $+3.6$ points. Averaged over all tests, \name reaches 52.8\%, compared with 50.3\% for vLLM-Incremental and 45.4\% for vLLM-Omni. Questions dropped while the server is busy count as incorrect. (Parentheses are percentage points change over the more accurate baseline.)}
\label{tab:online-accuracy}
\end{table*}

\begin{table*}[!t]
  \centering\small
  \setlength{\tabcolsep}{1pt}
  \renewcommand{\arraystretch}{1.18}
  \begin{tabularx}{\linewidth}{@{}cc*{6}{>{\centering\arraybackslash}X}@{}}
    \toprule
    \multirow{2}{*}{\textbf{Benchmark}} & \multirow{2}{*}{\textbf{Method}} & \textbf{Qwen3.5} & \textbf{Qwen3.5} & \mbox{\textbf{Nemotron-VL}} & \textbf{Qwen3.5} & \textbf{Qwen3.6} & \textbf{Qwen3.5} \\
    & & \textbf{0.8B} & \textbf{9B} & \textbf{Nano-12B} & \textbf{27B} & \textbf{35B-A3B} & \textbf{122B-A10B} \\
    \midrule
    \multirow{3}{*}{\textbf{PhoStream}} & vLLM-Omni & 41.59 & 60.87 & 50.66 & 69.00 & \textbf{69.65} & \textbf{72.14} \\
    & vLLM-Incremental & 41.59 & 62.57 & \textbf{51.61} & 68.05 & \textbf{69.65} & 71.14 \\
    \rowcolor{LightSteelBlue!20}\cellcolor{white} & \textcolor{SteelBlue!50!black}{\textbf{\name (Ours)}} & \textbf{41.97}\,{\scriptsize$(+0.4\%)$} & \textbf{63.33}\,{\scriptsize$(+0.8\%)$} & 51.04\,{\scriptsize$(-0.6\%)$} & \textbf{69.94}\,{\scriptsize$(+0.9\%)$} & 68.66\,{\scriptsize$(-1.0\%)$} & 71.14\,{\scriptsize$(-1.0\%)$} \\
    \addlinespace[5pt]
    \multirow{3}{*}{\textbf{StreamingBench}} & vLLM-Omni & 72.32 & 83.64 & 81.41 & 86.06 & 80.63 & 86.88 \\
    & vLLM-Incremental & \textbf{73.13} & \textbf{85.66} & \textbf{83.43} & \textbf{88.08} & \textbf{83.13} & \textbf{88.13} \\
    \rowcolor{LightSteelBlue!20}\cellcolor{white} & \textcolor{SteelBlue!50!black}{\textbf{\name (Ours)}} & 72.12\,{\scriptsize$(-1.0\%)$} & 85.25\,{\scriptsize$(-0.4\%)$} & 81.82\,{\scriptsize$(-1.6\%)$} & \textbf{88.08}\,{\scriptsize$(+0.0\%)$} & 82.50\,{\scriptsize$(-0.6\%)$} & 86.88\,{\scriptsize$(-1.3\%)$} \\
    \addlinespace[5pt]
    \multirow{3}{*}{\textbf{FastBench}} & vLLM-Omni & 25.49 & 33.99 & 30.72 & 39.87 & 36.27 & 41.18 \\
    & vLLM-Incremental & \textbf{27.78} & \textbf{35.62} & 30.39 & 40.52 & \textbf{36.60} & \textbf{42.16} \\
    \rowcolor{LightSteelBlue!20}\cellcolor{white} & \textcolor{SteelBlue!50!black}{\textbf{\name (Ours)}} & \textbf{27.78}\,{\scriptsize$(+0.0\%)$} & \textbf{35.62}\,{\scriptsize$(+0.0\%)$} & \textbf{31.37}\,{\scriptsize$(+0.7\%)$} & \textbf{41.83}\,{\scriptsize$(+1.3\%)$} & 35.95\,{\scriptsize$(-0.7\%)$} & \textbf{42.16}\,{\scriptsize$(+0.0\%)$} \\
    \addlinespace[5pt]
    \multirow{3}{*}{\textbf{BEYOND3D}} & vLLM-Omni & 28.76 & 37.90 & 34.58 & 49.14 & \textbf{37.97} & 37.77 \\
    & vLLM-Incremental & 25.56 & \textbf{38.18} & 34.19 & \textbf{49.23} & 36.47 & 36.45 \\
    \rowcolor{LightSteelBlue!20}\cellcolor{white} & \textcolor{SteelBlue!50!black}{\textbf{\name (Ours)}} & \textbf{30.58}\,{\scriptsize$(+1.8\%)$} & 37.47\,{\scriptsize$(-0.7\%)$} & \textbf{34.77}\,{\scriptsize$(+0.2\%)$} & 48.04\,{\scriptsize$(-1.2\%)$} & 37.26\,{\scriptsize$(-0.7\%)$} & \textbf{38.82}\,{\scriptsize$(+1.0\%)$} \\
    \bottomrule
  \end{tabularx}
  \caption{\textbf{Offline accuracy (\%, $\uparrow$).} \name's windowed state maintenance preserves the backbone's answer quality: without timing pressure, it stays within $-1.6$ to $+1.8$ points of the more accurate baseline in each of the 24 tests and has the highest mean accuracy, 54.4\% vs.\ 54.1\% for vLLM-Incremental and 53.7\% for vLLM-Omni. Combined with the large online gains (Table~\ref{tab:online-accuracy}), this shows that \name's online advantage comes from answering on time, not from changing what the model sees. (Parentheses are percentage points change over the more accurate baseline.)}
  \label{tab:offline}
\end{table*}

\subsection{Experimental Setup}
\label{sec:setup}

\textbf{Models.} We evaluate \name on six hybrid VLMs to assess scalability across model sizes and applicability across recurrent architectures.

\textit{Gated Delta Net~(GDN)}. We use the Qwen3.5~\cite{qwen35} and Qwen3.6 families, whose hybrid backbones combine GDN with attention layers. We evaluate three dense variants (0.8B, 9B, and 27B) and two mixture-of-experts~(MoE) variants (Qwen3.6-35B-A3B and Qwen3.5-122B-A10B) to examine how \name's latency benefits and memory overhead vary across model sizes and configurations.

\textit{State Space Model~(SSM)}. We use Nemotron-Nano-12B-v2-VL~\cite{nemotronnanovl}, whose hybrid backbone combines attention with Mamba-2~\cite{mamba2} state-space layers. This model allows us to assess whether \name's approach to context window maintenance extends beyond GDN-based architectures.

\textbf{Benchmarks.} We evaluate \name on four complementary benchmarks covering general video understanding, long-context reasoning, rapidly changing scenes, and spatial memory relevant to embodied applications. For online evaluation (\S\ref{sec:eval-online}), we use the first five hours of each benchmark. For offline evaluation (\S\ref{sec:eval-offline}), we use the full benchmarks, except for the 35B and 122B models, which use the first five hours due to limited GPU availability.

\begin{enumerate}[label=(\arabic*)]
  \item \textit{StreamingBench}~\cite{streamingbench} evaluates streaming video understanding through multiple-choice questions. We use it to establish whether \name preserves the backbone model's video understanding accuracy.

  \item \textit{PhoStream}~\cite{phostream} extends the evaluation to open-ended question answering over long video streams in mobile scenarios. It assesses whether \name preserves the contextual information needed for temporal reasoning and free-form answer generation.

  \item \textit{FastBench}~\cite{fastbench} focuses on rapidly changing scenes with brief events that require dense temporal sampling. We use it to evaluate \name's latency and answer quality when capturing fine-grained temporal evidence requires frequent frame updates.

  \item \textit{Beyond3D}~\cite{beyond3d} evaluates spatiotemporal reasoning in egocentric videos, with questions about objects that have been moved and subsequently left the field of view. It tests whether \name preserves the evolving spatial context needed for embodied applications such as robotic assistance.
\end{enumerate}

\textbf{Baselines.} For each model and benchmark, we compare \name against two inference paths provided by vLLM-Omni~\cite{vllmomni}, an open-source inference engine for multimodal workloads.
\begin{enumerate}[label=(\arabic*)]
  \item \textit{vLLM-Omni}: The native per-query inference path, which prefills the video context when a query arrives.

  \item \textit{vLLM-Omni + Incremental Prefill}~\cite{vllmomni}: The streaming inference path, which prefills incoming frames ahead of queries and reuses the resulting prefix cache at query time.\end{enumerate}

\textbf{Metrics.} We evaluate \name in terms of inference efficiency and answer quality. For efficiency, we report time-to-first-token~(TTFT), measured from query arrival to the first output token. For answer quality, we report results in both online and offline settings for each model--benchmark pair, using accuracy for multiple-choice tasks and benchmark-specific quality scores for open-ended tasks.

\textbf{Hardware.} In the online experiments, we use the smallest GPU available to us for each model. This is an A100 40\,GB for Qwen3.5-9B and Nemotron-VL Nano-12B, a GH200 for Qwen3.5-27B, a B200 for Qwen3.6-35B-A3B, and a B300 for Qwen3.5-122B-A10B. The memory ablation (Table~\ref{tab:memory-ablation}) runs each model on one A100 40\,GB GPU.

\subsection{Online Serving}
\label{sec:eval-online}

In the online setting, benchmark videos stream in real time. A client sends frames at wall-clock speed, with questions arriving at scheduled timestamps while frames continue to buffer. This tests a method's accuracy and responsiveness under live compute and demand constraints. The server processes one question at a time following the given question time in the benchmark. If the system is currently busy, new questions are dropped rather than queued, penalizing slow methods with lost questions. Incoming frames, however, are never dropped and are buffered.

\textbf{Takeaways.} \name sharply shortens the latency tail (Table~\ref{tab:online-p95}): it lowers p95 TTFT in all 20 tests by a geometric mean of \TTFTSpeedup over the faster baseline (up to $35.7\times$) and $33.8\times$ over vLLM-Omni, keeping p95 under 1\,s in 16 of 20 tests versus 3 for incremental prefill. Because slow answers drop questions, this also raises accuracy (Table~\ref{tab:online-accuracy}): \name averages 52.8\% versus 50.3\% for incremental prefill, gaining up to 18.4 points on Beyond3D. Appendix~\ref{app:online-p50} reports p50 TTFT (Table~\ref{tab:online-p50}) and full TTFT CDFs (Figure~\ref{fig:all-ttft-cdfs}).

\subsection{Offline Answer Quality}
\label{sec:eval-offline}

In the offline setting, the client sends all frames up to a question's timestamp as fast as the server can ingest them, then pauses for the server's answer before continuing. The server applies the same bounded frame window as in the online setting. This removes wall-clock constraints to isolate answer quality from streaming speed, ensuring accuracy differences stem solely from how each policy represents the video.

\textbf{Takeaways.} Without timing pressure, \name matches baseline answer quality (Table~\ref{tab:offline}): it stays within $-1.6$ to $+1.8$ points of the more accurate baseline in each of the 24 tests and has the highest mean accuracy, 54.4\% versus 54.1\% and 53.7\%. Its small offline Beyond3D differences, against large online gains, indicate that the online advantage comes from answering on time.

\subsection{Component Ablations}

\begin{figure}
  \centering
  \includegraphics[width=\columnwidth]{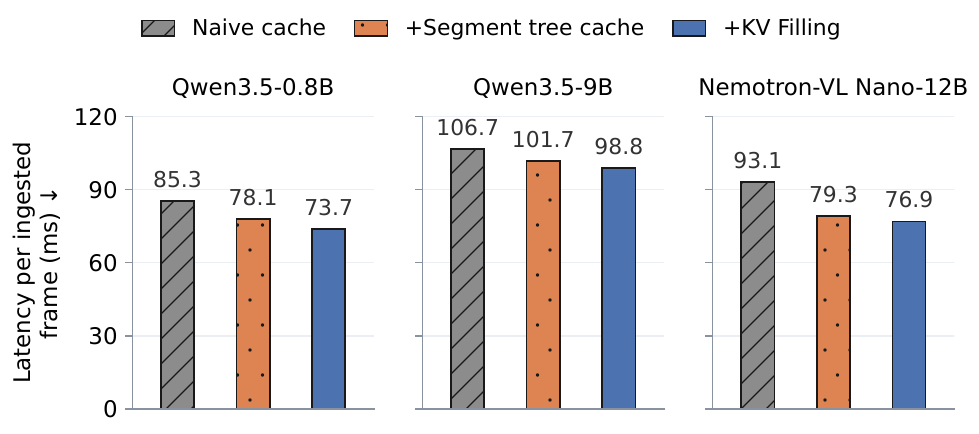}
  \caption{Component ablation: latency per ingested frame (ms, $\downarrow$) as caching components are added, from a naive cache to the segment tree cache and KV filling.}
  \label{fig:component-latency}
\end{figure}

\begin{table}
  \centering
  \small
  \setlength{\tabcolsep}{2.5pt}
  \renewcommand{\arraystretch}{1.15}
  \begin{tabularx}{0.9\columnwidth}{@{}c*{2}{>{\centering\arraybackslash}X}c@{}}
    \toprule
    \multirow{2}{*}{Configuration} & \multicolumn{2}{c}{Qwen3.5} & Nemotron-VL \\
    \cmidrule(lr){2-3}\cmidrule(l){4-4}
    & 0.8B & 9B & Nano-12B \\
    \midrule
    No \name              &  5.53 & 27.33 & 27.06 \\
    + Naive cache         &  8.52 & 30.93 & 31.06 \\
    + Segment tree        & 10.77 & 33.93 & 35.43 \\
    + KV Filling          &  9.43 & 32.16 & 35.05 \\
    \bottomrule
  \end{tabularx}
  \caption{\textbf{Peak GPU memory (GiB, $\downarrow$).} Each of \name's components has a modest memory cost: the segment tree trades 2.3--4.4\,GiB for 4.7--14.8\% lower per-frame latency (Figure~\ref{fig:component-latency}), and KV filling recovers 0.38--1.77\,GiB of it while lowering latency further. Overall, the full system uses 9.2\% more memory on average than the naive cache.}
  \label{tab:memory-ablation}
  \vspace{5mm}
\end{table}

We expect the segment tree to reduce recurrent-state reconstruction work at the cost of extra cached summaries (\S\ref{sec:summary-tree}), and KV filling to reduce attention-cache reclamation work (\S\ref{sec:windowed-attention-state}). We compare naive caching, the segment tree cache, and the segment tree with KV filling on Qwen3.5-0.8B, Qwen3.5-9B, and Nemotron-VL Nano-12B, using the same settings as the PhoStream evaluation with Qwen3.5. Figure~\ref{fig:component-latency} reports latency per ingested frame; Table~\ref{tab:memory-ablation} reports peak allocated GPU memory during ingestion and question answering, including a no-caching reference.

\textbf{Frame ingestion.} Adding the segment tree lowers latency by 4.7--14.8\% relative to naive caching, with the largest reduction on Nano-12B, from 93.1 to 79.3\,ms per frame. KV filling, added with the segment tree already enabled, lowers it by a further 2.9--5.6\%, for a total reduction of 7.4--17.4\%.

\textbf{Memory cost.} Table~\ref{tab:memory-ablation} shows that compared to naive caching the segment tree adds 2.25--4.37\,GiB of peak allocation. Adding KV filling reduces peak allocation by 0.38--1.77\,GiB relative to the segment tree alone. Overall, the net overhead remains modest: peak allocation increases by 9.2\% on average across the three models relative to naive caching.

\subsection{Fidelity under Window Expiry}

\begin{figure}
  \centering
  \begin{subfigure}[t]{0.48\columnwidth}
    \centering
    \includegraphics[width=\linewidth]{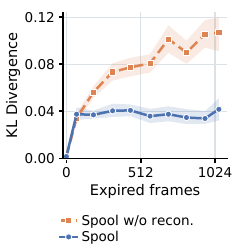}
    \caption{KL divergence.}
    \label{fig:logit-kl}
  \end{subfigure}
  \hfill
  \begin{subfigure}[t]{0.48\columnwidth}
    \centering
    \includegraphics[width=\linewidth]{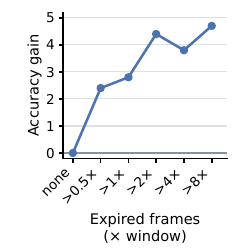}
    \caption{Accuracy gain.}
    \label{fig:acc-gap}
  \end{subfigure}
\caption{\textbf{Effect of state reconstruction on output fidelity.}
\textbf{(a)} Output KL divergence from vLLM-Omni remains near 0.04 through 1024 expired frames with reconstruction, but accumulates without it.
\textbf{(b)} Accuracy gain (points) from reconstruction on PhoStream's backward questions, which target earlier video content, grouped by expired frames relative to window size.
Both variants perform identically before expiration; reconstruction gains 2.4 points after half a window expires and 3.8--4.7 points beyond two windows.}
  \vspace{5mm}
  \label{fig:reconstruction-fidelity}
\end{figure}

In Figure~\ref{fig:reconstruction-fidelity}(\subref{fig:logit-kl}), we plot the KL divergence relative to vLLM-Omni. We observe that, without continually updating the GDN state, error continues to accumulate over time. In contrast, with state reconstruction, the KL divergence stays roughly constant at about 0.04 as the window advances, showing that \name's approximation error is bounded rather than accumulating over time. We see this constant approximation across a variety of models and benchmarks in Appendix \ref{app:gdn-state-error}.

In Figure~\ref{fig:reconstruction-fidelity}(\subref{fig:acc-gap}), we plot PhoStream's "backward" questions, which ask about earlier parts of the video. We compare Spool with and without state reconstruction on the same questions. The two give identical answers when no frames have expired, but once old frames start dropping out, Spool's lead grows to about 4\textendash{}5 points. This suggests state reconstruction matters most when a question depends on context the window has already discarded. Appendix~\ref{app:accuracy-analysis} shows formally that when expired history hurts the current question, \name without state reconstruction is strictly less accurate than \name.

\section{Related Work}
\label{sec:related}

\textbf{Reusing model state.} SGLang and Marconi reuse matching prefixes, with Marconi also caching recurrent checkpoints~\citep{sglang,marconi}. Tail-Replay approximates matched-prefix state from a short tail of cached attention outputs~\citep{tailreplay}, while CacheBlend selectively recomputes tokens to reuse non-prefix KV chunks~\citep{cacheblend}. A recurrent checkpoint still incorporates expired blocks when the window's left boundary advances. \name instead composes all retained blocks' cached transformations from the fixed prefix anchor. This preserves the full retained window's recorded updates without replaying token activations.

\textbf{Recurrent composition and window aggregation.} Affine composition underlies efficient Mamba and delta-rule kernels~\citep{mamba,mamba2,deltanet,gdn}. ReplaySSM caches recurrent inputs for decoding and speculative suffix rollback~\citep{replayssm}. HLA-WM also caches affine transformations without training, but recomposes history selected by camera geometry for video generation~\citep{hlawm}. Associative window aggregation is established in stream processing, including Reactive Aggregator's trees and DABA's constant-time FIFO operations~\citep{reactiveagg,daba}. \name instead maintains the recent video window by caching interval compositions in a segment tree as the oldest blocks expire.

\textbf{Streaming video inference.} StreamingVLM and InfiniteVL adapt training or attention structure for long inputs~\citep{streamingvlm,infinitevl}, while Video DeltaNet targets video generation with hybrid recurrent attention~\citep{videodeltanet}. Training-free methods compress or retrieve video KV~\citep{hermes,rekv,streamkv}; SimpleStream demonstrates competitive recent-frame windows~\citep{simplestream}. KV retention alone leaves recurrent history unchanged. \name reconstructs recurrent state after expiry on a pretrained hybrid backbone without further training.

\textbf{Memory selection and storage.} StreamingLLM, H$_2$O, and SnapKV retain sink, recent, or important KV entries~\citep{streamingllm,h2o,snapkv}. DeltaS and GateSight use recurrent-state signals for KV selection~\citep{deltas,gatesight}, while DAMP compresses recurrent states with mixed precision~\citep{damp}. PagedAttention and OPKV manage allocation and recall~\citep{pagedattention,opkv}. \name instead reconstructs recurrent state after expiry, excluding expired blocks' recorded updates from each layer.

\section{Conclusion}
\label{sec:conclusion}

Hybrid VLMs need to expire old frames without replaying retained context.
\name does so without retraining through cached affine transformations, segment tree reconstruction, and KV tail filling.
Across five VLMs and four benchmarks, it achieves the lowest p95 TTFT in all 20 model--benchmark tests, with a geometric mean speedup of \TTFTSpeedup, and an average online accuracy gain of \OnlineGain, each against the stronger baseline for that metric.

\section*{Acknowledgements}
This work was supported by Quanta Computer, Inc. under the AIR Project and by NSF Award 2616029.

\section*{AI Usage Disclosure}
We used ChatGPT and Claude to help polish the manuscript text and assist with scripts used in this work. Our use of these tools was limited to writing and scripting assistance.

\appendix

\section*{Appendix}

\section{Monoid Structure of Recurrent State Summaries}
\label{app:summary-monoids}

\begin{theorem}[Monoid of affine summaries]
  \label{thm:affine-monoid}
  Affine summaries $(A, B)$ form a non-commutative monoid under the merge in Equation~\eqref{eq:summary-composition}, with identity $(I, 0)$.
\end{theorem}

\begin{proof}[Proof of Theorem~\ref{thm:affine-monoid}]
  Fix $d_k, d_v \geq 1$ and use exact arithmetic. Let $\mathcal{M}_A = \mathbb{R}^{d_k \times d_k} \times \mathbb{R}^{d_v \times d_k}$ be the space of affine summaries $(A, B)$. Denote the merge in Equation~\eqref{eq:summary-composition} by $\star$. Let $x_{a,k} = (A_{a,k}, B_{a,k})$ and $x_{k,b} = (A_{k,b}, B_{k,b})$ be arbitrary elements of $\mathcal{M}_A$, labeled by adjacent half-open block ranges $[a, k)$ and $[k, b)$ with $a < k < b$. Their chronological merge is
  \[
    \begin{aligned}
      x_{a,k} \star x_{k,b} &= (A_{a,b}, B_{a,b}) \\
      A_{a,b} &= A_{a,k} A_{k,b} \\
      B_{a,b} &= B_{a,k} A_{k,b} + B_{k,b}
    \end{aligned}
  \]
  The product $A_{a,k} A_{k,b}$ has shape $d_k \times d_k$, and the sum $B_{a,k} A_{k,b} + B_{k,b}$ has shape $d_v \times d_k$. Thus $\star$ is closed on $\mathcal{M}_A$.

  To prove associativity, introduce an arbitrary third summary $x_{b,c} = (A_{b,c}, B_{b,c})$ labeled by the next range $[b, c)$, with $b < c$. Expanding either parenthesization of $x_{a,k} \star x_{k,b} \star x_{b,c}$ using distributivity and associativity of matrix multiplication yields the same coordinates:
  \[
    \begin{aligned}
      A_{a,c} &= A_{a,k} A_{k,b} A_{b,c} \\
      B_{a,c} &= B_{a,k} A_{k,b} A_{b,c} \\
      &\quad + B_{k,b} A_{b,c} + B_{b,c}
    \end{aligned}
  \]
  Hence $\star$ is associative. Let $I$ be the $d_k \times d_k$ identity and $0$ the $d_v \times d_k$ zero matrix. For any $(A, B) \in \mathcal{M}_A$,
  \[
    \begin{aligned}
      (I, 0) \star (A, B) &= (I A, 0 A + B) = (A, B) \\
      (A, B) \star (I, 0) &= (A I, B I + 0) = (A, B)
    \end{aligned}
  \]
  Therefore $(I, 0)$ is a two-sided identity, and $(\mathcal{M}_A, \star)$ is a monoid.

  The monoid is not commutative. Choose any nonzero $E \in \mathbb{R}^{d_v \times d_k}$ and let $x = (2 I, 0)$ and $y = (I, E)$. Then
  \[
    x \star y = (2 I, E) \neq (2 I, 2 E) = y \star x \qedhere
  \]
\end{proof}

\begin{corollary}[Monoid of residual summaries]
  \label{cor:residual-monoid}
  Residual summaries $(D, B)$ form a non-commutative monoid under the merge in Equation~\eqref{eq:residual-composition}, with identity $(0, 0)$.
\end{corollary}

\begin{proof}[Proof of Corollary~\ref{cor:residual-monoid}]
  Let $(\mathcal{M}_A, \star)$ be the affine summary monoid of Theorem~\ref{thm:affine-monoid} and $\mathcal{M}_D = \mathbb{R}^{d_k \times d_k} \times \mathbb{R}^{d_v \times d_k}$ the space of residual summaries $(D, B)$. Denote the merge in Equation~\eqref{eq:residual-composition} by $\diamond$. The map $\Phi: \mathcal{M}_A \to \mathcal{M}_D$ given by $\Phi(A, B) = (A - I, B)$ is bijective, with inverse $\Phi^{-1}(D, B) = (I + D, B)$. Take arbitrary affine summaries $x_{a,k} = (A_{a,k}, B_{a,k})$ and $x_{k,b} = (A_{k,b}, B_{k,b})$ for adjacent ranges $[a, k)$ and $[k, b)$. Write $D_{a,k} = A_{a,k} - I$ and $D_{k,b} = A_{k,b} - I$. The residual merge is
  \[
    \begin{gathered}
      (D_{a,k}, B_{a,k}) \diamond (D_{k,b}, B_{k,b}) \\
      = (D_{a,b}, B_{a,b}) \\
      D_{a,b} = D_{a,k} + D_{k,b} + D_{a,k} D_{k,b} \\
      B_{a,b} = B_{a,k} + B_{a,k} D_{k,b} + B_{k,b}
    \end{gathered}
  \]
  Its coordinates satisfy
  \[
    \begin{aligned}
      D_{a,b} &= (I + D_{a,k}) (I + D_{k,b}) - I \\
      &= A_{a,k} A_{k,b} - I \\
      B_{a,b} &= B_{a,k} (I + D_{k,b}) + B_{k,b} \\
      &= B_{a,k} A_{k,b} + B_{k,b}
    \end{aligned}
  \]
  Consequently,
  \[
    \Phi(x_{a,k} \star x_{k,b}) = \Phi(x_{a,k}) \diamond \Phi(x_{k,b})
  \]
  The bijection therefore transports closure and associativity of $\star$ to $\diamond$, and transports the identity to $\Phi(I, 0) = (0, 0)$, where the two zero matrices have shapes $d_k \times d_k$ and $d_v \times d_k$. This proves that $(\mathcal{M}_D, \diamond)$ is a monoid and that $\Phi$ is an isomorphism. The residual monoid inherits non-commutativity from the affine monoid through this isomorphism.

  Finally, for any state $S \in \mathbb{R}^{d_v \times d_k}$,
  \[
    S A + B = S + S (A - I) + B
  \]
  Thus $\Phi$ preserves the represented state transformation. In either representation, merging the summaries of $[a, k)$ and $[k, b)$ applies $F_{a,k}$ first and $F_{k,b}$ second, as required by chronological composition.
\end{proof}

\section{Analysis of \name}
\label{app:accuracy-analysis}

In the following analysis, we prove that if the old history has a negative influence on the current question, then \name without state reconstruction is strictly worse than \name.

For $\xi=(H,X,q)$ with expired history $H$, retained window $X$, and query $q$, let $y=f(X,q)$ be the target among $K\ge2$ candidates and $C_T,C_F$ the full \name and \name without state reconstruction caches before the query.

Write $z_{\xi,j}(C)\in\mathbb R$ for the full-model candidate score, with argmax prediction and fixed tie-breaking. Set $D_\xi=C_F-C_T$ and $C_\xi(t)=C_T+tD_\xi$, $0\le t\le1$, and define
\begin{align*}
  g_{\xi,j}(C)&=z_{\xi,y}(C)-z_{\xi,j}(C),\qquad j\ne y,\\
  h_{\xi,j}(t)&=g_{\xi,j}(C_\xi(t)).
\end{align*}
Assume the scores are $C^2$ near these paths; all bounds below are uniform over an input class $\mathcal A$.

\begin{theorem}[Margins and accuracy]\label{thm:margin-improvement}
  Suppose $a>r/2\ge0$ and, for every $\xi\in\mathcal A$, $j\ne y$, and $t\in[0,1]$,
  \[
    h'_{\xi,j}(0)\le-a,\qquad h''_{\xi,j}(t)\le r.
  \]
  Set $\Delta=a-r/2$ and $m_F(\xi)=\min_{j\ne y}g_{\xi,j}(C_F)$. Then
  \begin{equation}\label{eq:margin-improvement}
    g_{\xi,j}(C_T)\ge g_{\xi,j}(C_F)+\Delta.
  \end{equation}
  For any distribution supported on $\mathcal A$,
  \[
    \operatorname{Acc}(T)-\operatorname{Acc}(F)\ge\Pr(-\Delta<m_F(\xi)<0).
  \]
\end{theorem}

\begin{proof}
  For $h=h_{\xi,j}$, Taylor's theorem gives
  \[
    h(1)-h(0)\le h'(0)+r/2\le-\Delta.
  \]
  Since $\min_{j\ne y}g_{\xi,j}(C_T)\ge m_F+\Delta$, correctness of \name without state reconstruction implies correctness of \name, and only \name is correct on $-\Delta<m_F<0$. Take expectations.
\end{proof}

\begin{corollary}[Sampling and perplexity]\label{cor:softmax}
  For untruncated candidate softmax at temperature $\tau>0$, write $p_A=p_A(y\mid\xi)$ and $\ell_A=-\log p_A$. Under Theorem~\ref{thm:margin-improvement}, with $c=e^{-\Delta/\tau}$,
  \begin{equation}\label{eq:probability}
    p_T\ge\frac{p_F}{p_F+c(1-p_F)}>p_F.
  \end{equation}
  If $0<l\le p_F\le u<1$, then
  \begin{align*}
    p_T-p_F&\ge(1-c)l(1-u),\\
    \ell_T-\ell_F&\le\log\bigl(u+c(1-u)\bigr)<0,\\
    \frac{\operatorname{PPL}_T}{\operatorname{PPL}_F}&\le u+c(1-u)<1,
  \end{align*}
  where $\operatorname{PPL}_A=\exp[n^{-1}\sum_{i=1}^n\ell_A(\xi_i)]$ for $n\ge1$ evaluation instances in $\mathcal A$. Token perplexity uses the full vocabulary, $\tau=1$, and common teacher-forced prefixes.
\end{corollary}

\begin{proof}
  Equation~\eqref{eq:margin-improvement} implies
  \[
    p_T^{-1}-1=\sum_{j\ne y}e^{-g_{\xi,j}(C_T)/\tau}\le c(p_F^{-1}-1).
  \]
  Rearranging proves~\eqref{eq:probability}. Since $p_F+c(1-p_F)\le1$, the probability gap is at least $(1-c)p_F(1-p_F)$. Taking logs gives $\ell_T-\ell_F\le\log[p_F+c(1-p_F)]$. Apply the probability bounds, then average and exponentiate for perplexity.
\end{proof}

\paragraph{Recurrent layers.} For recurrent updates affine in state at fixed layer inputs, record \name without state reconstruction's maps $S_i=S_{i-1}A_i^F+B_i^F$, $i=1,\ldots,n$. Starting from $S_{\rm left}^F$ or prefix state $S_P$ gives $S_F$ or $S_{F0}$, respectively. Subtraction yields
\begin{equation}\label{eq:recurrent-history}
  S_F-S_{F0}=(S_{\rm left}^F-S_P)A_1^F\cdots A_n^F.
\end{equation}
Form $C_{F0}$ by replacing \name without state reconstruction's recurrent states with $S_{F0}$. Set $D_0=C_F-C_{F0}$ and $E=C_{F0}-C_T$, so $D_\xi=D_0+E$. Bounds
\begin{align*}
  \nabla g_{\xi,j}(C_T)^\top D_0&\le-a_0,\\
  \nabla g_{\xi,j}(C_T)^\top E&\le b_0,
\end{align*}
for every $j\ne y$ give $h'_{\xi,j}(0)\le-a_0+b_0$ by linearity. Theorem~\ref{thm:margin-improvement} applies with $a=a_0-b_0$ under the same curvature bound and $a_0-b_0>r/2$.

\section{GDN State Error}
\label{app:gdn-state-error}

We compare \name with \name without reconstruction on Beyond3D dataset on Qwen3.5 series. Both methods consume identical encoded visual features and are compared against vLLM for each question.

\begin{figure*}
  \centering
  \begin{subfigure}[t]{0.295\textwidth}
    \centering
    \includegraphics[width=\linewidth]{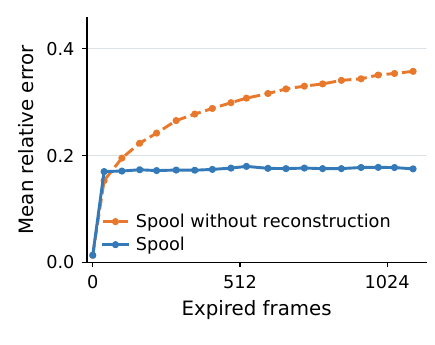}
    \caption{Qwen3.5-0.8B.}
    \label{fig:gdn-state-error-08b}
  \end{subfigure}
  \begin{subfigure}[t]{0.295\textwidth}
    \centering
    \includegraphics[width=\linewidth]{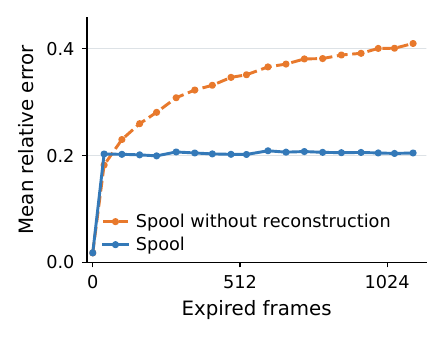}
    \caption{Qwen3.5-9B.}
    \label{fig:gdn-state-error-9b}
  \end{subfigure}
  \begin{subfigure}[t]{0.295\textwidth}
    \centering
    \includegraphics[width=\linewidth]{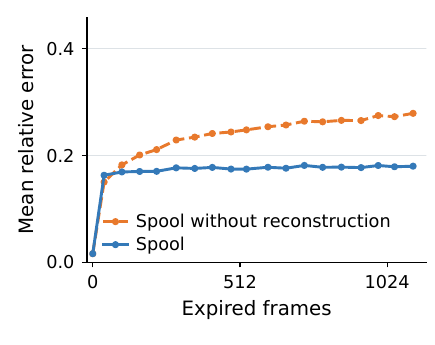}
    \caption{Qwen3.5-27B.}
    \label{fig:gdn-state-error-27b}
  \end{subfigure}
  \caption{\textbf{Mean GDN state error.} Relative Frobenius error averaged over each model's GDN layers, comparing \name with \name without reconstruction against the vLLM reference. The evaluation uses all 9,000 Beyond3D questions.}
  \label{fig:gdn-state-error}
\end{figure*}

For method $M$, we report the mean relative state error
\begin{equation}
  \bar e_M=\frac{1}{L}\sum_{\ell=1}^{L}\frac{\lVert S_{\ell,q}^{M}-S_{\ell,q}^{\mathrm{vLLM}}\rVert_F}{\lVert S_{\ell,q}^{\mathrm{vLLM}}\rVert_F},
\end{equation}
where $L$ is the number of GDN layers. Figure~\ref{fig:gdn-state-error} shows mean relative error in GDN state matrix; horizontal positions are the mean expired-frame counts in each bin.

\subsection{Per-Layer Error in Qwen3.5-9B}
\label{app:gdn-state-error-9b-layerwise}

\begin{figure*}[p]
  \centering
  \captionsetup[subfigure]{font=scriptsize,skip=2pt}
  \includegraphics[width=\textwidth]{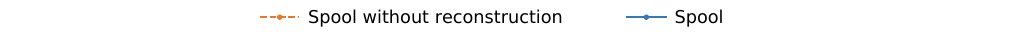}\par
  \smallskip
  \begin{subfigure}[t]{0.24\textwidth}
    \centering
    \includegraphics[width=\linewidth]{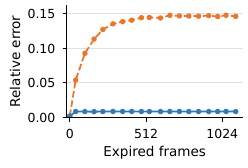}
    \caption{GDN 1}
    \label{fig:gdn-state-error-9b-layer-01}
  \end{subfigure}\hfill\begin{subfigure}[t]{0.24\textwidth}
    \centering
    \includegraphics[width=\linewidth]{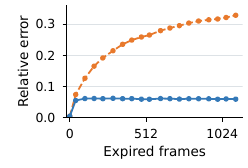}
    \caption{GDN 2}
    \label{fig:gdn-state-error-9b-layer-02}
  \end{subfigure}\hfill\begin{subfigure}[t]{0.24\textwidth}
    \centering
    \includegraphics[width=\linewidth]{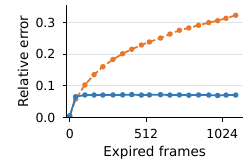}
    \caption{GDN 3}
    \label{fig:gdn-state-error-9b-layer-03}
  \end{subfigure}\hfill\begin{subfigure}[t]{0.24\textwidth}
    \centering
    \includegraphics[width=\linewidth]{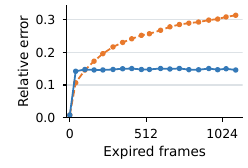}
    \caption{GDN 4}
    \label{fig:gdn-state-error-9b-layer-04}
  \end{subfigure}\par\smallskip
  \begin{subfigure}[t]{0.24\textwidth}
    \centering
    \includegraphics[width=\linewidth]{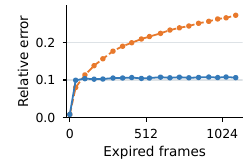}
    \caption{GDN 5}
    \label{fig:gdn-state-error-9b-layer-05}
  \end{subfigure}\hfill\begin{subfigure}[t]{0.24\textwidth}
    \centering
    \includegraphics[width=\linewidth]{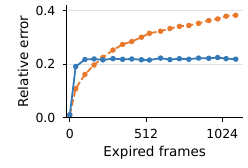}
    \caption{GDN 6}
    \label{fig:gdn-state-error-9b-layer-06}
  \end{subfigure}\hfill\begin{subfigure}[t]{0.24\textwidth}
    \centering
    \includegraphics[width=\linewidth]{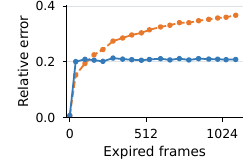}
    \caption{GDN 7}
    \label{fig:gdn-state-error-9b-layer-07}
  \end{subfigure}\hfill\begin{subfigure}[t]{0.24\textwidth}
    \centering
    \includegraphics[width=\linewidth]{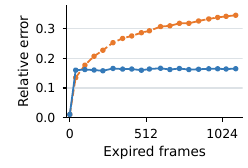}
    \caption{GDN 8}
    \label{fig:gdn-state-error-9b-layer-08}
  \end{subfigure}\par\smallskip
  \begin{subfigure}[t]{0.24\textwidth}
    \centering
    \includegraphics[width=\linewidth]{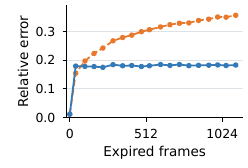}
    \caption{GDN 9}
    \label{fig:gdn-state-error-9b-layer-09}
  \end{subfigure}\hfill\begin{subfigure}[t]{0.24\textwidth}
    \centering
    \includegraphics[width=\linewidth]{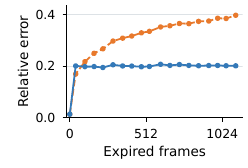}
    \caption{GDN 10}
    \label{fig:gdn-state-error-9b-layer-10}
  \end{subfigure}\hfill\begin{subfigure}[t]{0.24\textwidth}
    \centering
    \includegraphics[width=\linewidth]{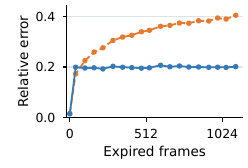}
    \caption{GDN 11}
    \label{fig:gdn-state-error-9b-layer-11}
  \end{subfigure}\hfill\begin{subfigure}[t]{0.24\textwidth}
    \centering
    \includegraphics[width=\linewidth]{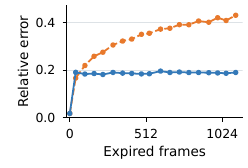}
    \caption{GDN 12}
    \label{fig:gdn-state-error-9b-layer-12}
  \end{subfigure}\par\smallskip
  \begin{subfigure}[t]{0.24\textwidth}
    \centering
    \includegraphics[width=\linewidth]{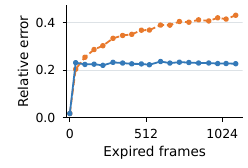}
    \caption{GDN 13}
    \label{fig:gdn-state-error-9b-layer-13}
  \end{subfigure}\hfill\begin{subfigure}[t]{0.24\textwidth}
    \centering
    \includegraphics[width=\linewidth]{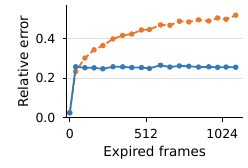}
    \caption{GDN 14}
    \label{fig:gdn-state-error-9b-layer-14}
  \end{subfigure}\hfill\begin{subfigure}[t]{0.24\textwidth}
    \centering
    \includegraphics[width=\linewidth]{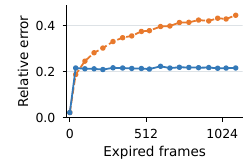}
    \caption{GDN 15}
    \label{fig:gdn-state-error-9b-layer-15}
  \end{subfigure}\hfill\begin{subfigure}[t]{0.24\textwidth}
    \centering
    \includegraphics[width=\linewidth]{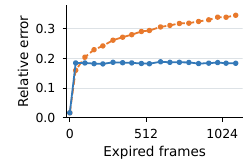}
    \caption{GDN 16}
    \label{fig:gdn-state-error-9b-layer-16}
  \end{subfigure}\par\smallskip
  \begin{subfigure}[t]{0.24\textwidth}
    \centering
    \includegraphics[width=\linewidth]{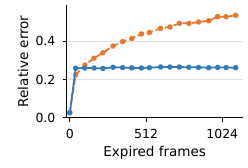}
    \caption{GDN 17}
    \label{fig:gdn-state-error-9b-layer-17}
  \end{subfigure}\hfill\begin{subfigure}[t]{0.24\textwidth}
    \centering
    \includegraphics[width=\linewidth]{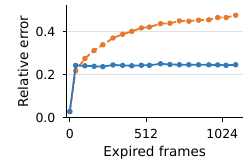}
    \caption{GDN 18}
    \label{fig:gdn-state-error-9b-layer-18}
  \end{subfigure}\hfill\begin{subfigure}[t]{0.24\textwidth}
    \centering
    \includegraphics[width=\linewidth]{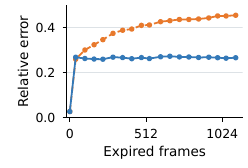}
    \caption{GDN 19}
    \label{fig:gdn-state-error-9b-layer-19}
  \end{subfigure}\hfill\begin{subfigure}[t]{0.24\textwidth}
    \centering
    \includegraphics[width=\linewidth]{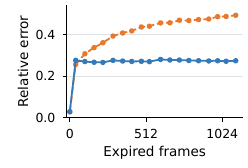}
    \caption{GDN 20}
    \label{fig:gdn-state-error-9b-layer-20}
  \end{subfigure}\par\smallskip
  \begin{subfigure}[t]{0.24\textwidth}
    \centering
    \includegraphics[width=\linewidth]{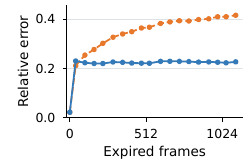}
    \caption{GDN 21}
    \label{fig:gdn-state-error-9b-layer-21}
  \end{subfigure}\hfill\begin{subfigure}[t]{0.24\textwidth}
    \centering
    \includegraphics[width=\linewidth]{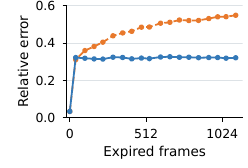}
    \caption{GDN 22}
    \label{fig:gdn-state-error-9b-layer-22}
  \end{subfigure}\hfill\begin{subfigure}[t]{0.24\textwidth}
    \centering
    \includegraphics[width=\linewidth]{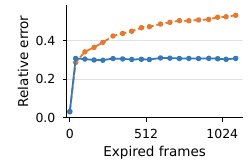}
    \caption{GDN 23}
    \label{fig:gdn-state-error-9b-layer-23}
  \end{subfigure}\hfill\begin{subfigure}[t]{0.24\textwidth}
    \centering
    \includegraphics[width=\linewidth]{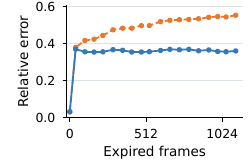}
    \caption{GDN 24}
    \label{fig:gdn-state-error-9b-layer-24}
  \end{subfigure}
  \caption{\textbf{Per-layer GDN state error for Qwen3.5-9B.} Relative Frobenius error against the vLLM reference for all 24 GDN layers, averaged over questions within each expired-frame bin. The evaluation uses all 9,000 Beyond3D questions and a 64-frame retained window.}
  \label{fig:gdn-state-error-9b-layerwise}
\end{figure*}

Figure~\ref{fig:gdn-state-error-9b-layerwise} reports the relative state error separately for all 24 GDN layers of Qwen3.5-9B under the same evaluation protocol.

\section{Additional Online Latency Results}
\label{app:online-p50}

Table~\ref{tab:online-p50} reports online p50 TTFT for the same configurations as Table~\ref{tab:online-p95}.

\begin{table*}[!t]
  \centering\small
  \setlength{\tabcolsep}{1pt}
  \renewcommand{\arraystretch}{1.18}
  \begin{tabularx}{\linewidth}{@{}cc*{5}{>{\centering\arraybackslash}X}@{}}
    \toprule
    \multirow{2}{*}{\textbf{Benchmark}} & \multirow{2}{*}{\textbf{Method}} & \textbf{Qwen3.5} & \mbox{\textbf{Nemotron-VL}} & \textbf{Qwen3.5} & \textbf{Qwen3.6} & \textbf{Qwen3.5} \\
    & & \textbf{9B} & \textbf{Nano-12B} & \textbf{27B} & \textbf{35B-A3B} & \textbf{122B-A10B} \\
    \midrule
    \multirow{3}{*}{\textbf{PhoStream}} & vLLM-Omni & 27.08 & 5.35 & 12.03 & 6.86 & 4.85 \\
    & vLLM-Incremental & 1.42 & 0.47 & \textbf{0.28} & 0.37 & 0.31 \\
    \rowcolor{LightSteelBlue!20}\cellcolor{white} & \textcolor{SteelBlue!50!black}{\textbf{\name (Ours)}} & \textbf{0.33}\,{\scriptsize$(\times4.3)$} & \textbf{0.15}\,{\scriptsize$(\times3.1)$} & \textbf{0.28}\,{\scriptsize$(\times1.0)$} & \textbf{0.19}\,{\scriptsize$(\times1.9)$} & \textbf{0.28}\,{\scriptsize$(\times1.1)$} \\
    \addlinespace[5pt]
    \multirow{3}{*}{\textbf{StreamingBench}} & vLLM-Omni & 28.39 & 5.44 & 9.59 & 7.93 & 5.16 \\
    & vLLM-Incremental & 0.78 & 0.42 & \textbf{0.25} & 0.30 & \textbf{0.28} \\
    \rowcolor{LightSteelBlue!20}\cellcolor{white} & \textcolor{SteelBlue!50!black}{\textbf{\name (Ours)}} & \textbf{0.33}\,{\scriptsize$(\times2.4)$} & \textbf{0.15}\,{\scriptsize$(\times2.8)$} & 0.27\,{\scriptsize$(\times0.9)$} & \textbf{0.20}\,{\scriptsize$(\times1.5)$} & 0.29\,{\scriptsize$(\times1.0)$} \\
    \addlinespace[5pt]
    \multirow{3}{*}{\textbf{FastBench}} & vLLM-Omni & 3.36 & 5.46 & 4.90 & 0.60 & 0.75 \\
    & vLLM-Incremental & 0.15 & 2.66 & 0.23 & 0.15 & 0.23 \\
    \rowcolor{LightSteelBlue!20}\cellcolor{white} & \textcolor{SteelBlue!50!black}{\textbf{\name (Ours)}} & \textbf{0.12}\,{\scriptsize$(\times1.3)$} & \textbf{0.14}\,{\scriptsize$(\times19.6)$} & \textbf{0.20}\,{\scriptsize$(\times1.1)$} & \textbf{0.12}\,{\scriptsize$(\times1.3)$} & \textbf{0.19}\,{\scriptsize$(\times1.2)$} \\
    \addlinespace[5pt]
    \multirow{3}{*}{\textbf{BEYOND3D}} & vLLM-Omni & 10.39 & 35.12 & 53.92 & 72.09 & 58.56 \\
    & vLLM-Incremental & 0.87 & 1.87 & 1.17 & 2.07 & 2.64 \\
    \rowcolor{LightSteelBlue!20}\cellcolor{white} & \textcolor{SteelBlue!50!black}{\textbf{\name (Ours)}} & \textbf{0.40}\,{\scriptsize$(\times2.1)$} & \textbf{0.46}\,{\scriptsize$(\times4.1)$} & \textbf{0.75}\,{\scriptsize$(\times1.6)$} & \textbf{0.91}\,{\scriptsize$(\times2.3)$} & \textbf{1.48}\,{\scriptsize$(\times1.8)$} \\
    \bottomrule
  \end{tabularx}
  \caption{\textbf{Online p50 TTFT (s, $\downarrow$).} \name's latency advantage is concentrated in the tail: vLLM-Incremental answers typical questions quickly but stalls on a minority, which dominates its p95 (Table~\ref{tab:online-p95}). \name still has the lowest or tied-lowest p50 in 18 of 20 tests ($1.9\times$ geometric mean over the faster baseline, $32.2\times$ over vLLM-Omni), with the largest gap on FastBench with Nemotron-VL Nano-12B (0.14\,s vs.\ 2.66\,s). vLLM-Incremental is marginally faster on StreamingBench with Qwen3.5-27B and Qwen3.5-122B-A10B, by 0.01--0.02\,s. Parentheses: speedup over the faster baseline.}
  \label{tab:online-p50}
\end{table*}

\begin{figure*}
  \centering
  \includegraphics[width=0.245\textwidth]{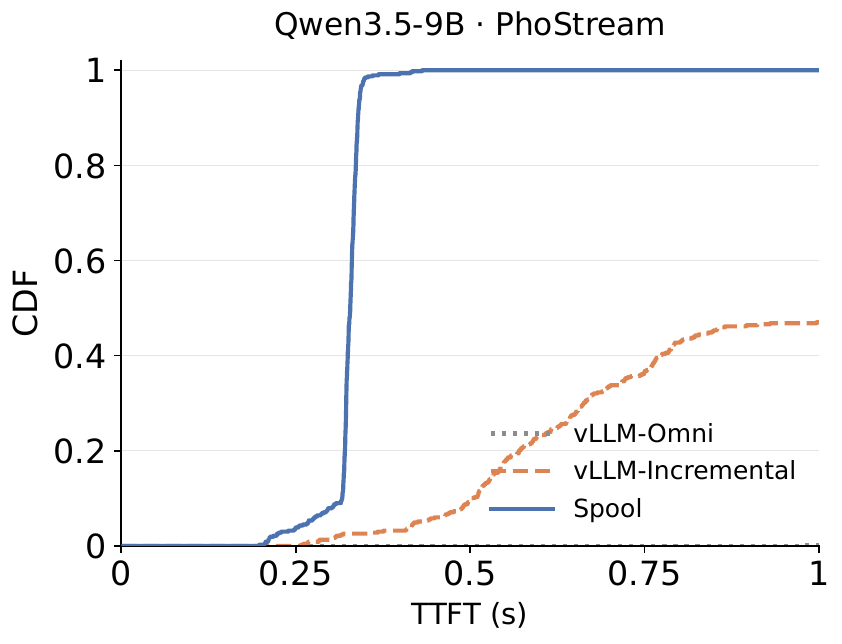}\hfill\includegraphics[width=0.245\textwidth]{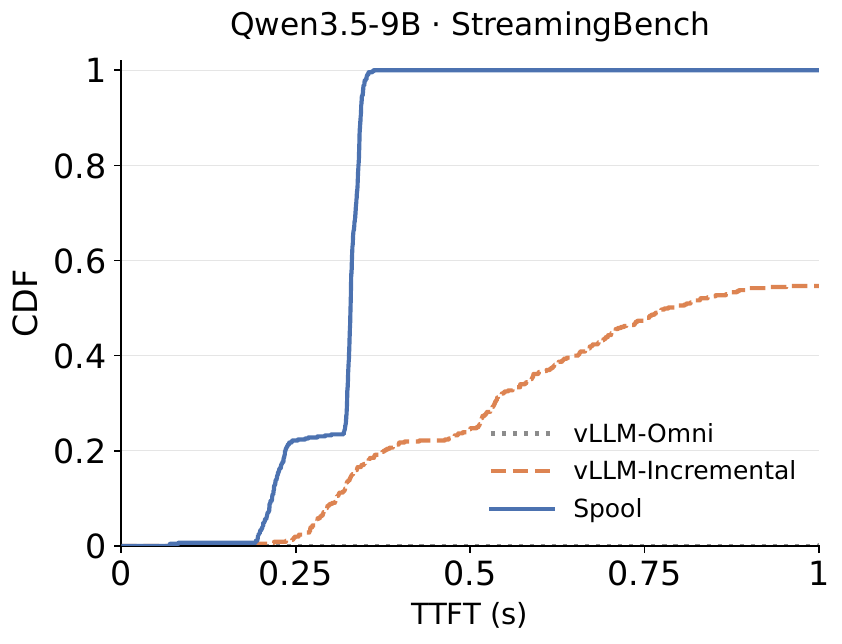}\hfill\includegraphics[width=0.245\textwidth]{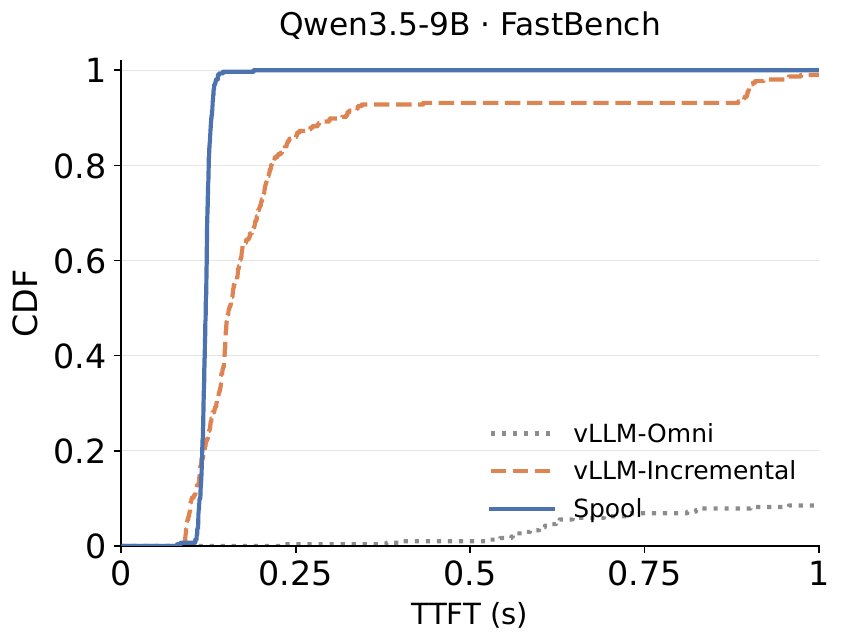}\hfill\includegraphics[width=0.245\textwidth]{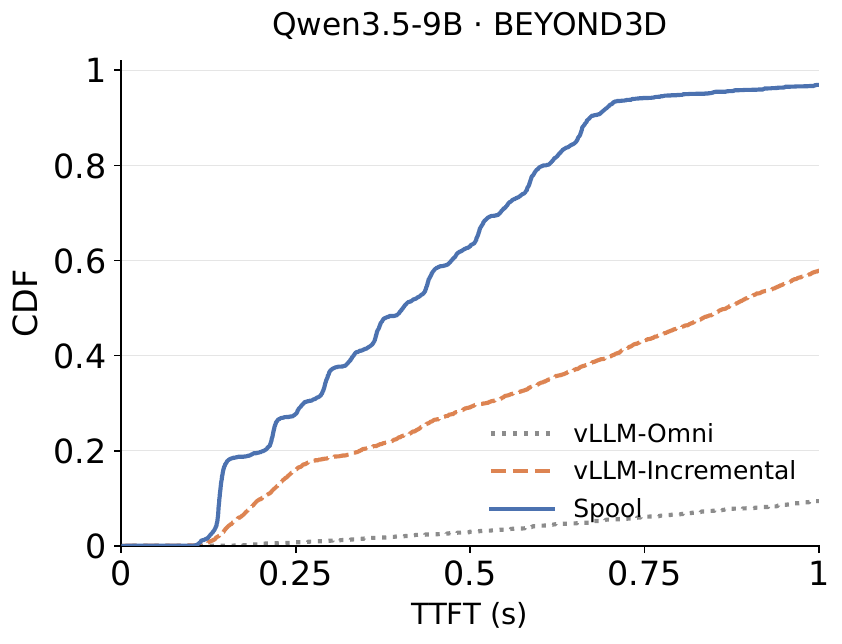}\par
  \includegraphics[width=0.245\textwidth]{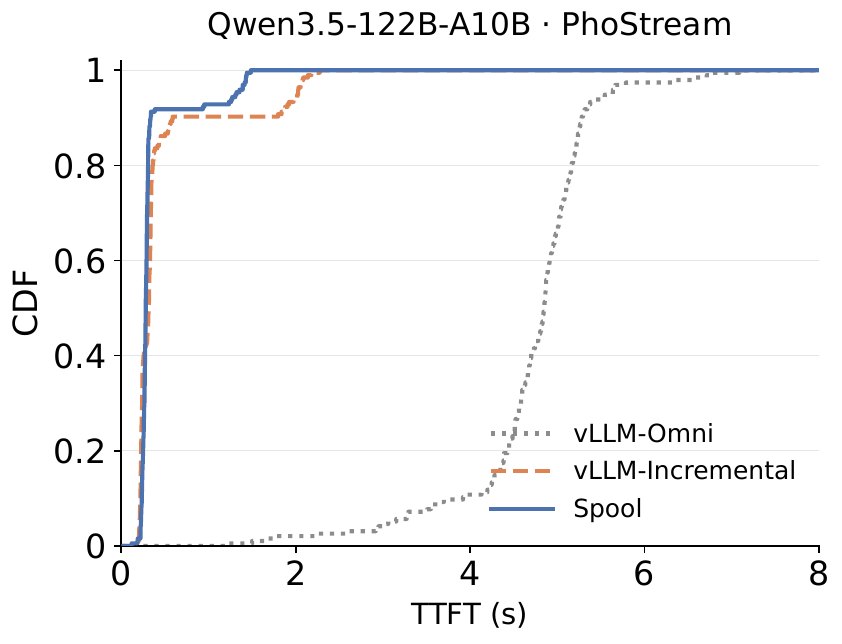}\hfill\includegraphics[width=0.245\textwidth]{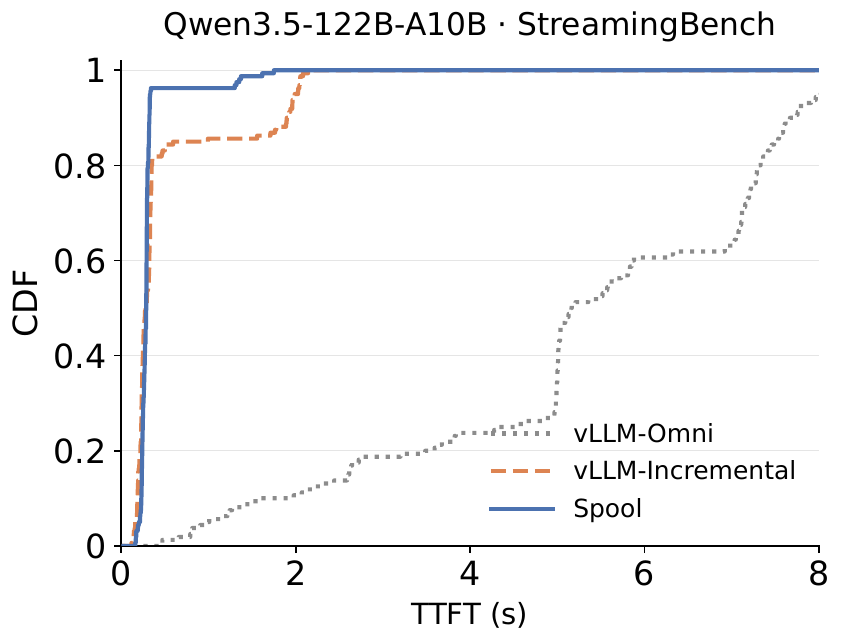}\hfill\includegraphics[width=0.245\textwidth]{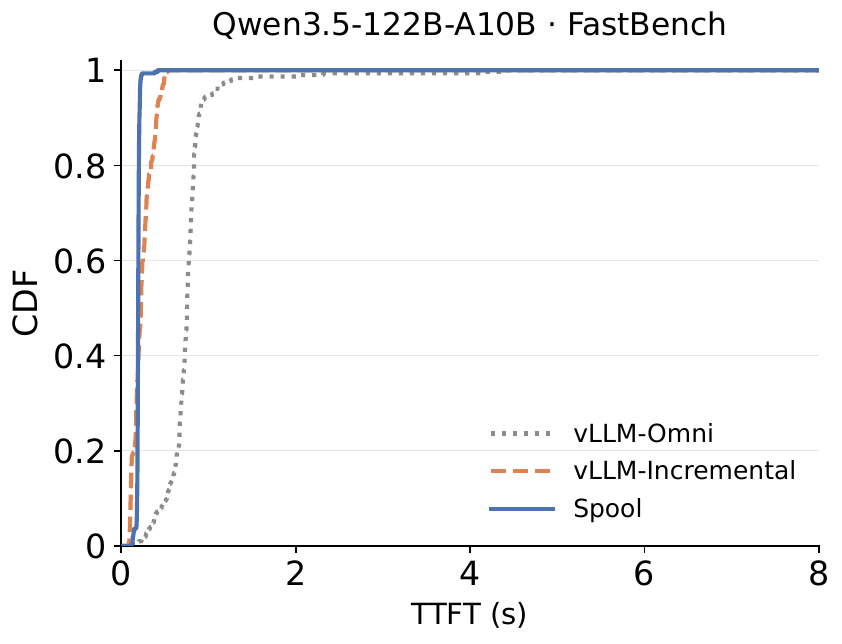}\hfill\includegraphics[width=0.245\textwidth]{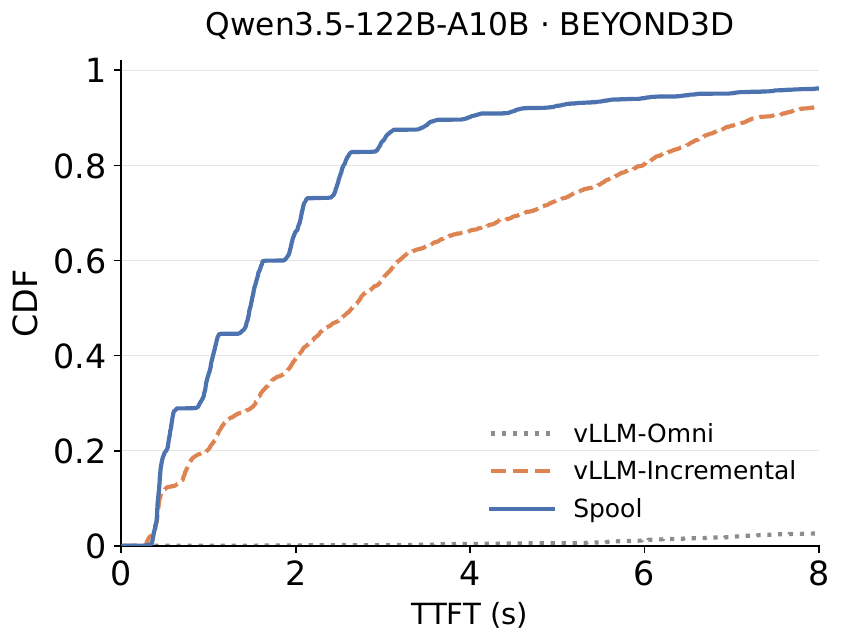}\par
  \includegraphics[width=0.245\textwidth]{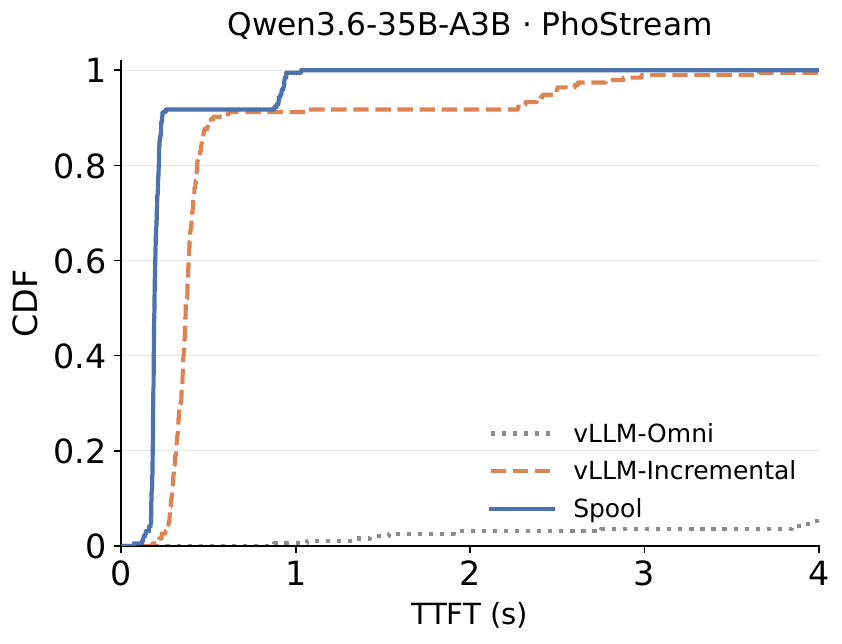}\hfill\includegraphics[width=0.245\textwidth]{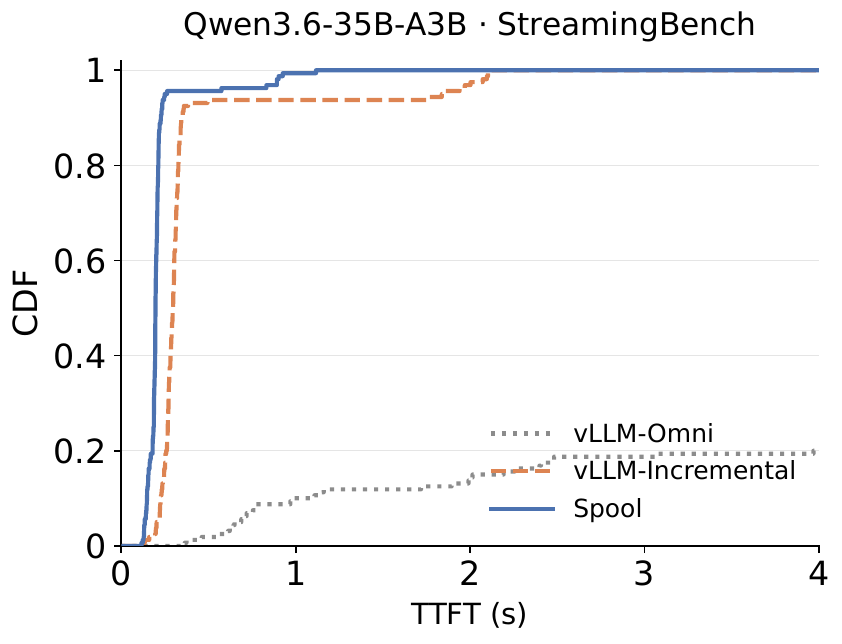}\hfill\includegraphics[width=0.245\textwidth]{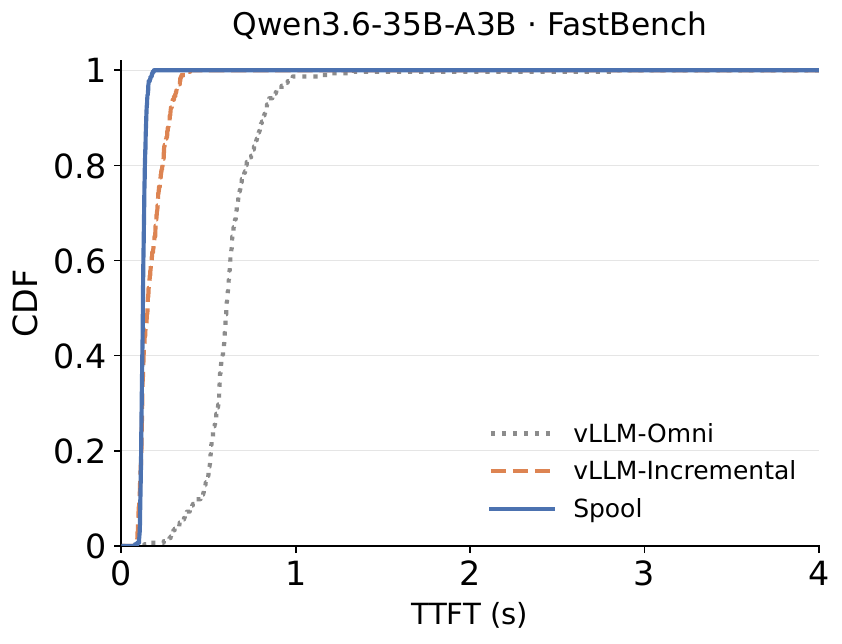}\hfill\includegraphics[width=0.245\textwidth]{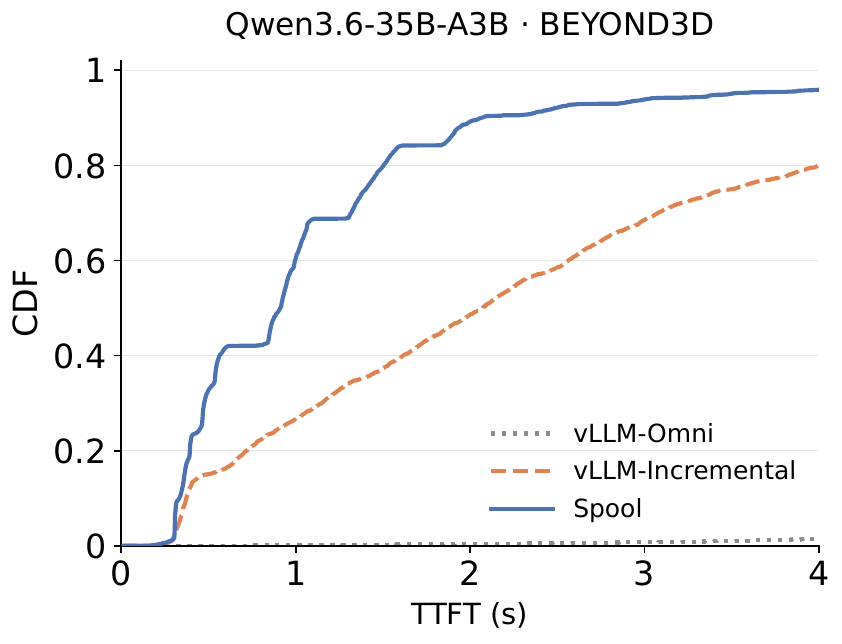}\par
  \includegraphics[width=0.245\textwidth]{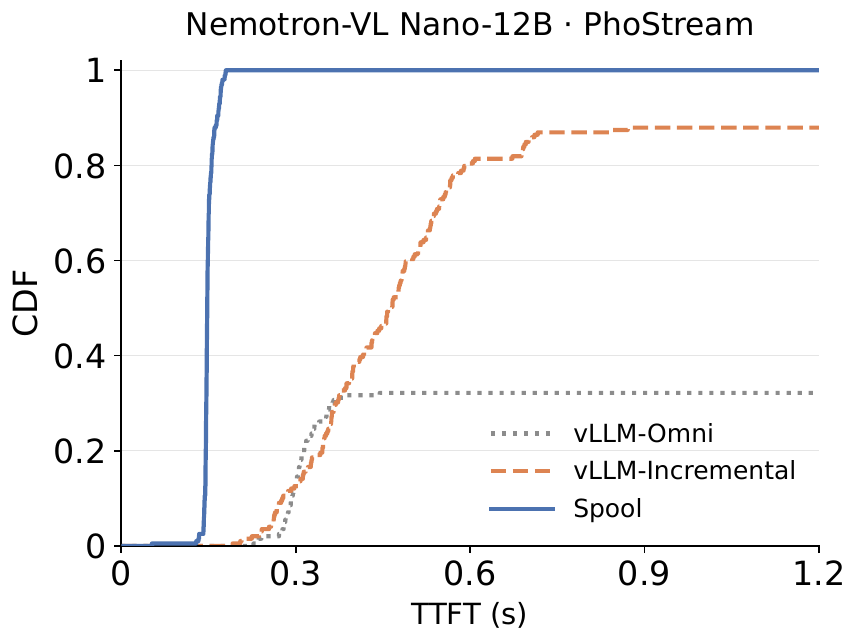}\hfill\includegraphics[width=0.245\textwidth]{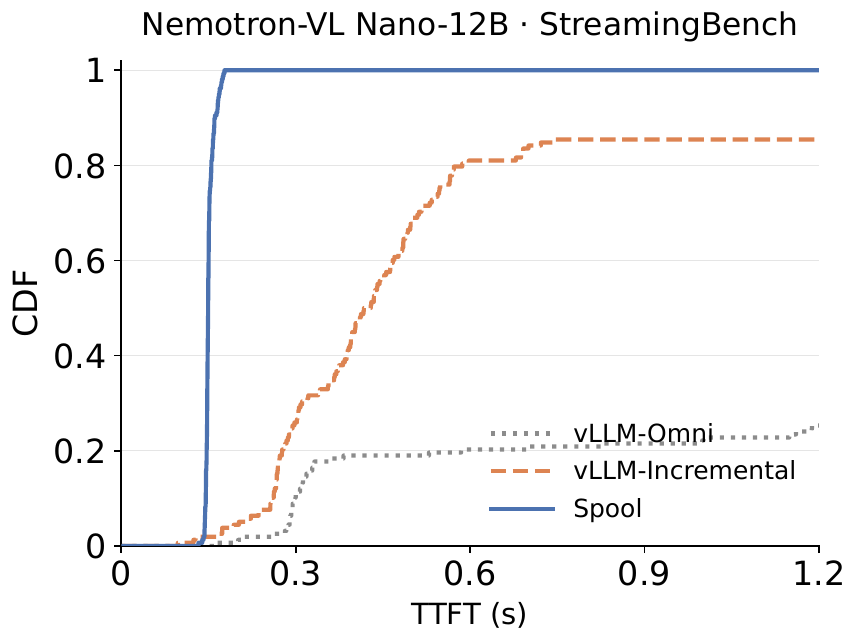}\hfill\includegraphics[width=0.245\textwidth]{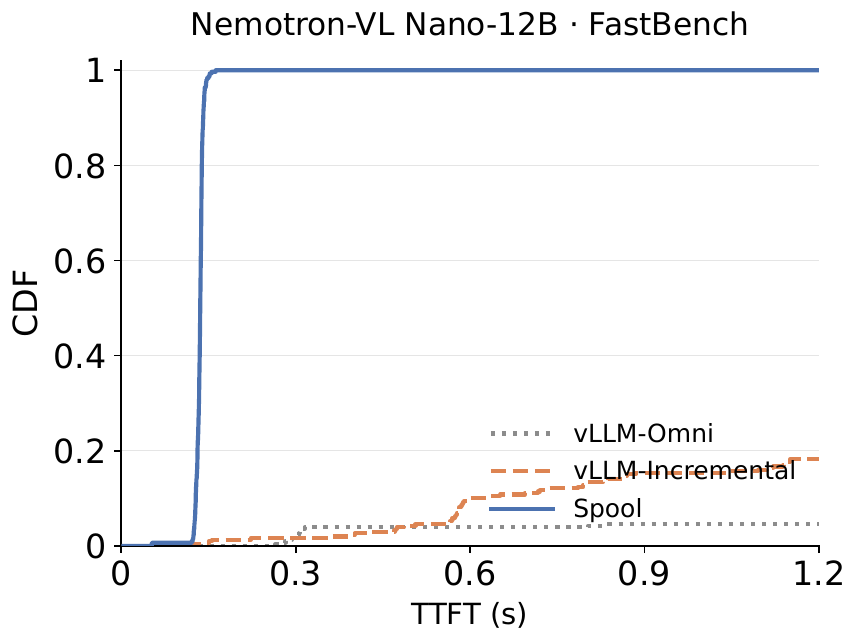}\hfill\includegraphics[width=0.245\textwidth]{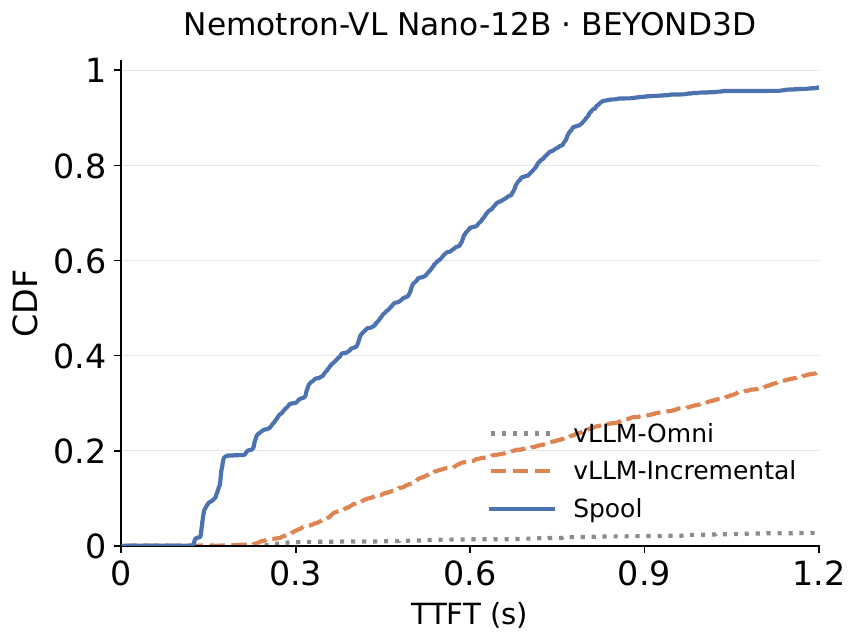}\par
  \includegraphics[width=0.245\textwidth]{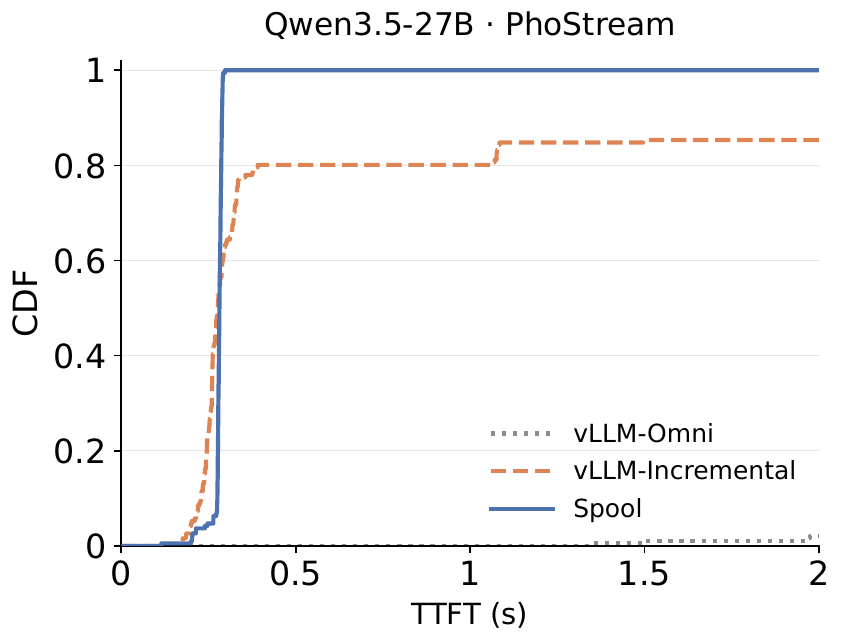}\hfill\includegraphics[width=0.245\textwidth]{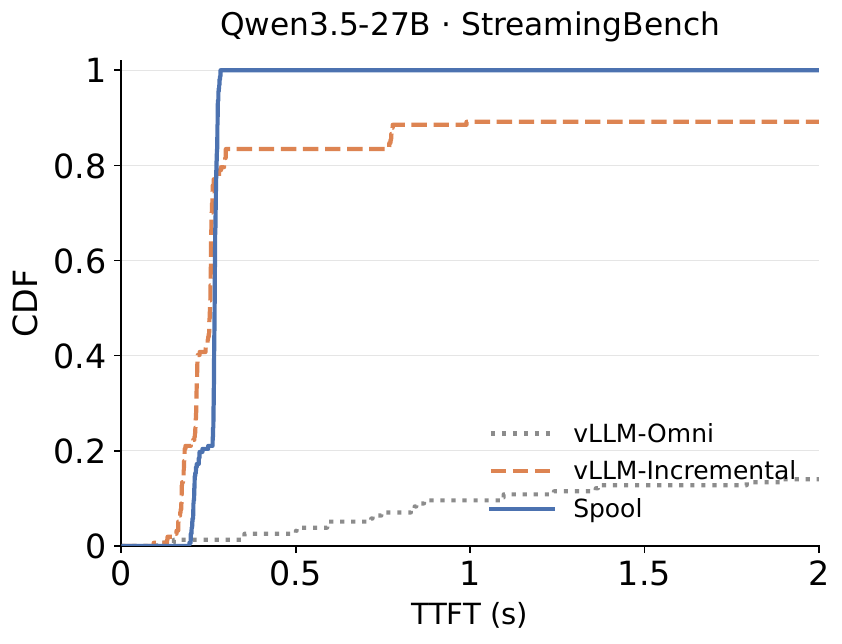}\hfill\includegraphics[width=0.245\textwidth]{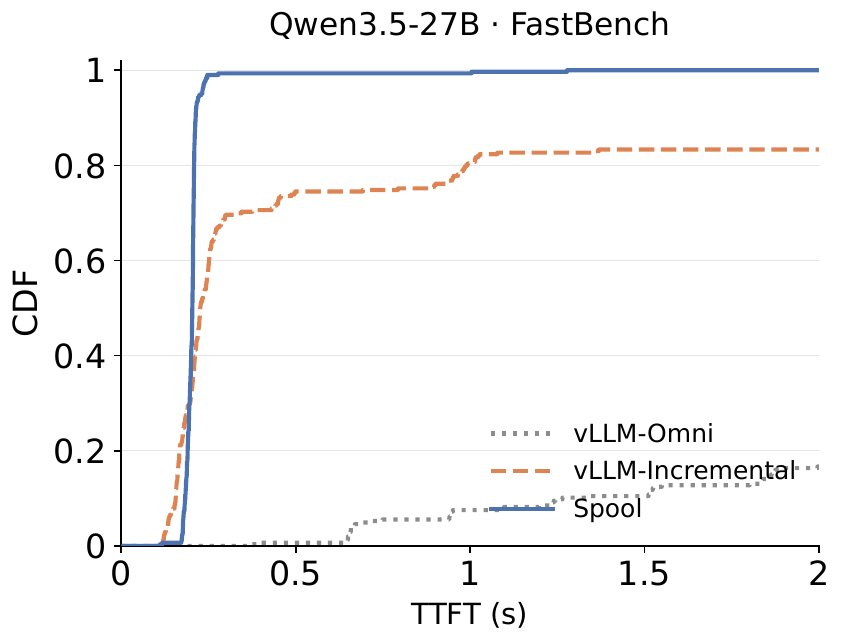}\hfill\includegraphics[width=0.245\textwidth]{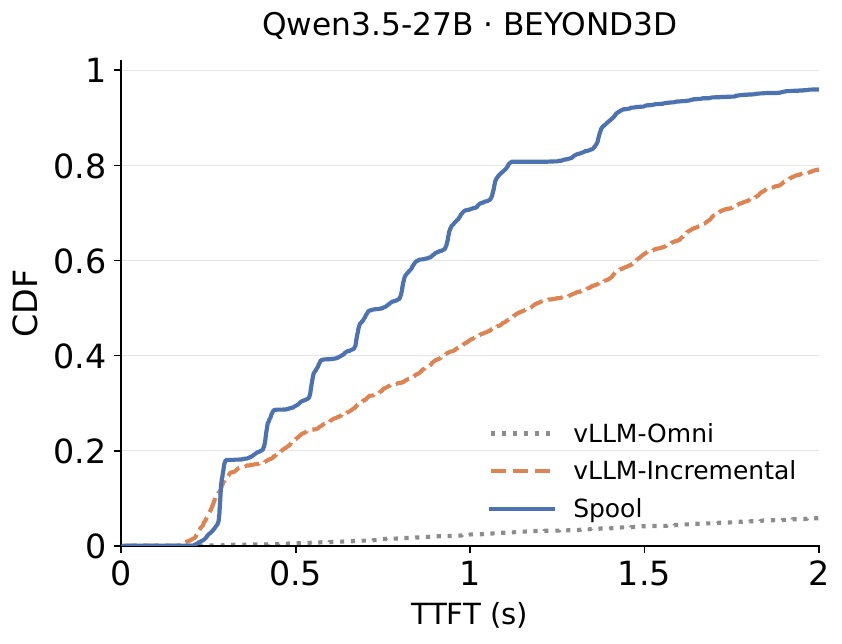}\par
  \caption{Empirical CDFs of online TTFT. Rows correspond to Qwen3.5-9B, Qwen3.5-122B-A10B, Qwen3.6-35B-A3B, Nemotron-VL Nano-12B, and Qwen3.5-27B. Columns correspond to PhoStream, StreamingBench, FastBench, and BEYOND3D.}
  \label{fig:all-ttft-cdfs}
\end{figure*}

\end{document}